\documentclass[lettersize,journal]{IEEEtran}
\usepackage{amsmath,amsfonts}
\usepackage{algorithmic}
\usepackage{algorithm}
\usepackage{array}
\usepackage[caption=false,font=normalsize,labelfont=sf,textfont=sf]{subfig}
\usepackage{textcomp}
\usepackage{stfloats}
\usepackage{url}
\usepackage{verbatim}
\usepackage{graphicx}
\usepackage{cite}
\usepackage{booktabs}
\usepackage{multirow}
\usepackage{amssymb} 
\usepackage{pifont}  
\usepackage{bm}  
\usepackage{rotating} 
\begin{document}

\title{Appearance Decomposition Gaussian Splatting for Multi-Traversal Reconstruction}

\author{Yangyi Xiao, Siting Zhu, Baoquan Yang, Tianchen Deng, Yongbo Chen~\IEEEmembership{Member,~IEEE}, Hesheng Wang~\IEEEmembership{Senior Member,~IEEE}
\thanks{\quad Corresponding Authors: Yongbo Chen, Hesheng Wang (email: \{shjtdx\_cyb, wanghesheng\}@sjtu.edu.cn)}
\thanks{\quad All authors are with Department of Automation, Key Laboratory of System Control and Information Processing of Ministry of Education, State Key Laboratory of Avionics Integration and Aviation System-of-Systems Synthesis, Shanghai Jiao Tong University, Shanghai 200240, China. (e-mail:\{x\_whyy, zhusiting, yangbaoquan, dengtianchen\}@sjtu.edu.cn )}

}


\markboth{Preprint}{Xiao \MakeLowercase{\textit{et al.}}: Appearance Decomposition Gaussian Splatting for Multi-Traversal Reconstruction}


\maketitle

\begin{abstract}
    Multi-traversal scene reconstruction is important for high-fidelity autonomous driving simulation and digital twin construction. This task involves integrating multiple sequences captured from the same geographical area at different times. In this context, a primary challenge is the significant appearance inconsistency across traversals caused by varying illumination and environmental conditions, despite the shared underlying geometry. This paper presents ADM-GS (Appearance Decomposition Gaussian Splatting for Multi-Traversal Reconstruction), a framework that applies an explicit appearance decomposition to the static background to alleviate appearance entanglement across traversals. For the static background, we decompose the appearance into traversal-invariant material, representing intrinsic material properties, and traversal-dependent illumination, capturing lighting variations. Specifically, we propose a neural light field that utilizes a frequency-separated hybrid encoding strategy. By incorporating surface normals and explicit reflection vectors, this design separately captures low-frequency diffuse illumination and high-frequency specular reflections. Quantitative evaluations on the Argoverse 2 and Waymo Open datasets demonstrate the effectiveness of ADM-GS. In multi-traversal experiments, our method achieves a +0.98 dB PSNR improvement over existing latent-based baselines while producing more consistent appearance across traversals. Code will be available at \url{https://github.com/IRMVLab/ADM-GS}.
\end{abstract}

\begin{IEEEkeywords}
Autonomous driving, Scene reconstruction, 3D Gaussian Splatting, Multi-traversal
\end{IEEEkeywords}
\section{Introduction}
    In the field of autonomous driving simulation and testing, constructing photorealistic, editable, and geometrically consistent digital twin scenes is important for validating systems from perception to planning. As the scale of data acquisition expands, the task of scene reconstruction is moving from single-sequence settings to large-scale multi-traversal scenarios. Compared with single-sequence approaches that are limited by fixed illumination conditions and inevitable geometric occlusions, multi-traversal integration provides more comprehensive viewpoint coverage to reduce geometric ambiguities while also capturing richer appearance variations across different times and conditions. However, multi-traversal reconstruction faces a central challenge: shared underlying geometry coexists with substantial appearance variation across traversals. While the physical layout, such as roads and buildings, remains largely consistent, appearance can vary significantly under different illumination conditions, times of day, and seasonal solar angles.
    
    Existing scene graph-based methods, including 4DGF~\cite{fischer2024dynamic} and MTGS~\cite{li2025mtgs}, typically decompose the scene into static backgrounds and dynamic objects, and employ environment latent codes or color residuals to model appearance differences across traversals. Although these methods have made significant progress in view synthesis, they still face limitations in multi-traversal settings. Latent-based approaches tend to entangle traversal-specific illumination effects with the static scene appearance. Under such frameworks, high-frequency lighting effects, including cast shadows and specular highlights, are often absorbed into the static appearance representation. This entanglement makes it difficult for the model to distinguish stable material appearance from traversal-dependent illumination, leading to artifacts under large cross-traversal lighting variations.
    
    Motivated by these limitations, we propose an appearance decomposition framework. Our key insight is that robust appearance modeling in multi-traversal scenes can be improved by explicitly decoupling traversal-invariant material from traversal-dependent illumination. However, incorporating such decomposition into a 3DGS framework~\cite{kerbl20233d} is non-trivial. First, the original 3DGS represents scenes with anisotropic ellipsoids for volumetric filling and lacks explicit surface normals, making normal-guided illumination estimation unstable. Second, the lighting environment in outdoor scenes is highly complex, containing both low-frequency diffuse effects and highly view-dependent high-frequency specular reflections on car bodies and glass surfaces. Without explicit frequency-aware guidance, a single illumination representation struggles to simultaneously model these distinct signals. Consequently, the network tends to conflate specular highlights with stable appearance, reducing consistency across traversals.

    \begin{figure*}[t]
      \centering
      \includegraphics[width=1.0\linewidth]{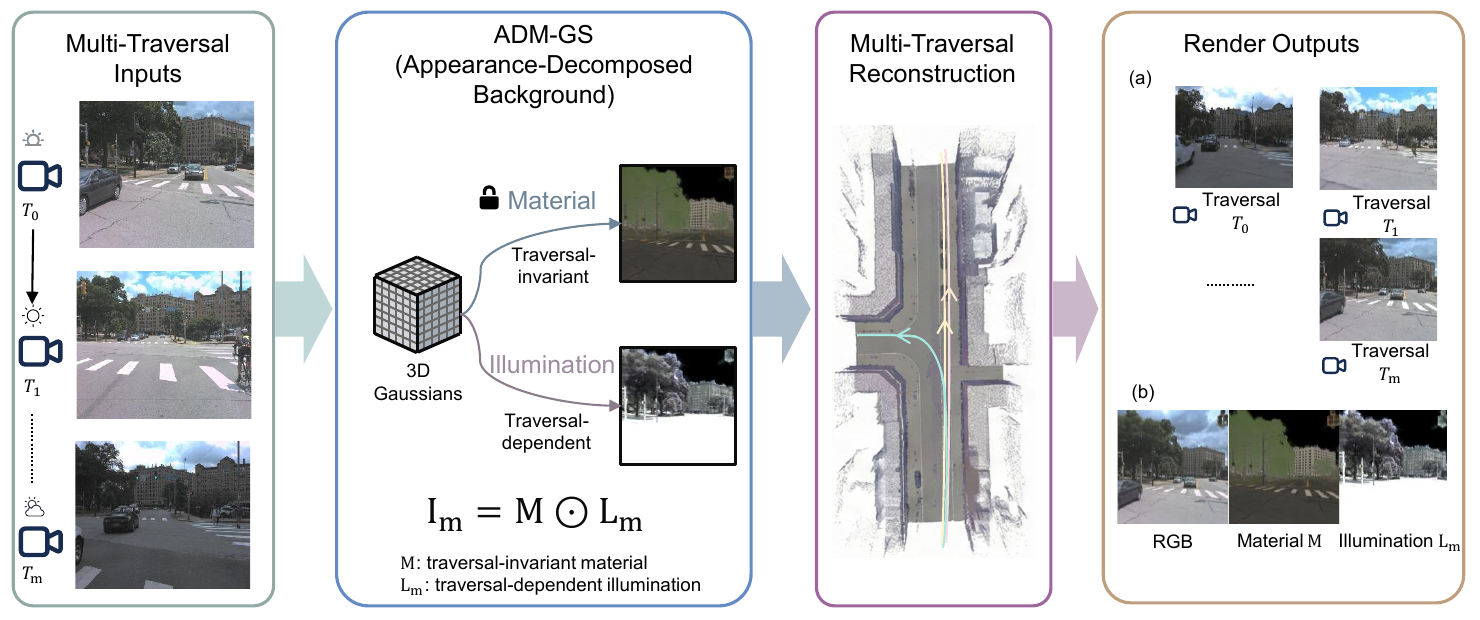} 
      \vspace{-0.5cm} 
      \caption{
        \textbf{Consistent Multi-Traversal Reconstruction via Appearance Decomposition.}
        \textbf{(Left)} Input images from multi-traversal scenes exhibit appearance inconsistencies due to changes in illumination, weather, and time of day.
        \textbf{(Middle)} ADM-GS decomposes the static scene appearance into a traversal-invariant material field ($M$) and a traversal-dependent light field ($L_m$), enabling structured modeling of cross-traversal appearance variation.
        \textbf{(Right)} Rendering results show that ADM-GS preserves consistent material across different environmental conditions while capturing traversal-specific illumination variations, leading to improved multi-traversal novel view synthesis.
      }
      \label{fig:teaser}
      \vspace{-0.3cm} 
    \end{figure*}

    Building on this formulation, we develop ADM-GS, a unified scene graph framework for multi-traversal reconstruction with explicit appearance decomposition. For the static background, ADM-GS decomposes appearance into traversal-invariant material and traversal-dependent illumination, enabling more structured modeling of cross-traversal appearance variation. To make this decomposition effective in 3DGS, we further introduce geometric regularization to stabilize surface-aware illumination estimation and incorporate reflection cues to better model view-dependent effects. In addition, a frequency-separated encoding strategy is designed to capture both low-frequency diffuse effects and high-frequency specular responses within a single light field. We further adopt transient geometry embeddings from 4DGF~\cite{fischer2024dynamic} to reduce the influence of geometrically inconsistent regions. In summary, the main contributions of this paper are as follows:
    
    \begin{itemize}
        \item We identify a key limitation of existing latent-based multi-traversal reconstruction methods, where traversal-specific illumination effects are often entangled with static scene appearance. To address this issue, we propose ADM-GS, an appearance decomposition framework that explicitly separates traversal-invariant material from traversal-dependent illumination.
    
        \item We develop a geometry-guided illumination modeling design for 3DGS, including surface-aware geometric regularization, explicit reflection vector injection, and frequency-separated encoding. These components stabilize illumination estimation and improve the modeling of both diffuse and view-dependent effects in complex outdoor scenes.
    
        \item Experiments on the Argoverse 2 and Waymo Open datasets demonstrate the effectiveness of our method in both single-traversal and multi-traversal reconstruction, with improved rendering quality and more consistent appearance across traversals.
    \end{itemize}

\section{Related Work}

\subsection{Autonomous Driving Scene Reconstruction}
    In computer vision and robotics, high-quality scene reconstruction~\cite{zhu2024sni, zhu2025sni} is important for simulation, scene understanding, and novel view synthesis. Recent surveys have also highlighted the growing role of neural and Gaussian-based scene representations in robotics and autonomous driving~\cite{deng2025best3dscenerepresentation, zhu20243d}. Related efforts have also started to extend Gaussian scene representations beyond reconstruction toward more general driving world models~\cite{deng2025gaussiandwm3dgaussiandriving}. The field has evolved from early implicit Neural Radiance Fields (NeRF)~\cite{mildenhall2021nerf} to recent explicit 3D Gaussian Splatting (3DGS)~\cite{kerbl20233d}. For large-scale dynamic urban scenarios, existing methods mainly follow two technical directions.
    
        \textit{Scene graph-based explicit decomposition:}
        To model dynamic elements, this line of work uses 3D bounding boxes to explicitly decompose the scene. Starting from Neural Scene Graphs (NSG)~\cite{ost2021neural}, subsequent works such as~\cite{xie2023snerf, yang2023unisim, tonderski2024neurad} incorporated multi-sensor fusion, including LiDAR and cameras, to improve geometric consistency and simulation quality. With the efficient rasterization of 3DGS, works such as~\cite{zhou2024drivinggaussian, yan2024street} transferred this explicit decomposition idea to Gaussian-based scene representations. To further reduce geometric artifacts under limited observations,~\cite{khan2025autosplat, zhou2024hugs, chen2025omnire} introduced priors such as symmetry constraints and motion models. More recent research has continued to improve scalability and multimodal fusion. For example,~\cite{hess2025splatad} optimizes trajectories via LiDAR-camera alignment, while~\cite{wang2025drivesplat} addresses detail balancing in large-scale environments. In addition,~\cite{wang2025unifying} introduced bilateral grids to improve pixel-level illumination alignment. Related work has also explored composite Gaussian representations for reconstruction and editing in dynamic autonomous driving scenes~\cite{xu2026drivingeditor}.
        
        \textit{Self-supervised dynamic modeling:}
        To reduce reliance on 3D annotations, another line of work explores self-supervised dynamic scene modeling. In implicit representations, \cite{turki2023suds, yangemernerf} used multi-field hash tables and flow fields to decouple scene dynamics without bounding boxes, while \cite{deng2023prosgnerf} introduced a progressive dynamic neural scene graph for large-scale urban scenes with fast-moving objects, and \cite{11127897} explored pose-free reconstruction. In the 3DGS setting, early attempts such as~\cite{huang2024s3gaussian, chen2023periodic} used HexPlane~\cite{cao2023hexplane} or periodic vibration functions to implicitly encode subtle motions. To improve geometric coherence, later methods such as~\cite{sun2025splatflow, song2025coda, peng2025desire} incorporated optical flow distillation and deformation compensation, while~\cite{xu2025ad} employed B-spline smoothing to constrain object trajectories. More recently, foundation-model-based 2D vision cues~\cite{ren2024grounded} have also been used for zero-shot instance decomposition, as in~\cite{zhang2025street, lindstrom2025idsplat}, which leverage semantic masks for controllable reconstruction without manual annotation.
            
    Although these methods have shown strong performance on single-sequence reconstruction, they are primarily designed for single-sequence settings. When extended to multi-traversal data with stronger appearance variation across time, they are not explicitly designed to model cross-traversal environmental changes, which can make consistent fusion more challenging.
    
\subsection{Appearance Modeling and Intrinsic Decomposition}
Robust appearance modeling is important for handling illumination and environmental variation in uncontrolled outdoor scenes. Existing methods mainly fall into two categories: latent-based appearance modeling and inverse rendering / intrinsic decomposition.

    \textit{Latent-based appearance modeling:}
    To address appearance discrepancies in unstructured photo collections, this line of work typically represents lighting variation with low-dimensional latent embeddings. NeRF-W~\cite{martin2021nerf} introduced appearance embeddings, followed by~\cite{chen2022hallucinated, li2023nerf}, which further improved robustness to occlusions and inter-sequence differences through appearance hallucination and triplet-based regularization. For city-scale scenes,~\cite{tancik2022block, turki2022mega} introduced exposure codes and spatially-aware embeddings to capture large-scale lighting variation. This paradigm was later adapted to 3DGS:~\cite{zhang2024gaussian, kulhanek2024wildgaussians} combine appearance codes with robust losses to handle transient disturbances, while~\cite{dahmani2024swag} introduces unsupervised appearance adjustment. For larger scenes,~\cite{lin2024vastgaussian} proposed a partitioned decoupling strategy. However, regardless of whether the representation takes the form of a global code, local feature, or bilateral grid, these methods usually model appearance variation implicitly, and may entangle traversal-dependent lighting effects with static scene appearance.
    
    \textit{Inverse rendering and intrinsic decomposition:}
    Another category of work explicitly incorporates intrinsic decomposition into neural rendering. In implicit neural fields,~\cite{zhang2021nerfactor, zhang2021physg} factorized the scene into intrinsic components such as normals, BRDFs, and illumination, enabling relighting and appearance editing. Subsequent works such as~\cite{jin2023tensoir, zhang2022modeling} further improved computational efficiency. For outdoor scenes,~\cite{rudnev2022nerf, sun2023sol} introduced explicit models for solar irradiance and skylight. With the rise of 3DGS, explicit primitives further accelerated this direction:~\cite{jiang2024gaussianshader} improved relighting quality with simplified illumination functions, while~\cite{gao2024relightable} incorporated ray tracing for more accurate shadow modeling. For large-scale urban scenes,~\cite{lin2025urbanir, chen2025invrgb+} introduced geometric priors and LiDAR intensity cues to reduce ambiguity beyond RGB-only inputs.

Although inverse rendering provides more structured appearance decomposition, many existing methods face a trade-off between physical detail, computational complexity, and large-scale scalability. Our work is closer in spirit to decomposition-based appearance modeling, but is designed for multi-traversal reconstruction under large-scale driving scenarios.

\subsection{Multi-Traversal Reconstruction}
Multi-traversal reconstruction aims to build scene representations with more complete geometry and broader view coverage by leveraging repeated observations of the same scene acquired at different times. To address appearance changes and transient dynamic elements, existing methods can be broadly divided into two streams: scene decomposition and continual adaptation, and structured scene graphs with appearance modeling.

    \textit{Scene decomposition and continual adaptation:}
    Early multi-traversal efforts mainly focused on reducing data heterogeneity through filtering or adaptation. Works such as~\cite{qin2024crowd, li2024memorize} use semantic segmentation or self-supervised decomposition to filter transient objects and align image styles across sources. To adapt to scene evolution over time,~\cite{zeng2025gaussianupdate} introduced continual learning into 3DGS through generative replay, while~\cite{han2025extrapolated} studied the limitations of current models in view extrapolation. However, these methods generally do not explicitly model the combination of shared static structure and traversal-specific appearance variation.
    
    \textit{Structured scene graphs with appearance modeling:}
    More recent multi-traversal methods adopt scene graph-based hierarchical modeling to represent shared static content together with traversal-specific dynamics. ML-NSG~\cite{fischer2024multi} introduced a multi-level neural scene graph with sequence-level latent codes, and 4DGF~\cite{fischer2024dynamic} extended this idea with dynamic 3D Gaussian fields and compact neural field fitting. HybridWorldSim~\cite{li2025hybridworldsim} further connected multi-traversal reconstruction with simulation by incorporating generative models for controllable rendering. MTGS~\cite{li2025mtgs} refined the graph structure and introduced color correction nodes and spherical harmonics (SH) residuals to compensate for illumination shifts.

Although methods such as ML-NSG~\cite{fischer2024multi}, 4DGF~\cite{fischer2024dynamic}, and MTGS~\cite{li2025mtgs} have improved multi-traversal geometric fusion through structured designs, their appearance modeling remains largely implicit. Whether using latent codes or SH residuals, these methods typically fit RGB observations directly, which can make it harder to separate traversal-dependent illumination changes from stable scene appearance. Our work addresses this issue by introducing explicit appearance decomposition into the multi-traversal reconstruction framework, with the goal of improving cross-traversal consistency while remaining compatible with large-scale driving scenes.


\section{METHOD}
    For large scale multi-traversal autonomous driving scene reconstruction, our core objective is to handle cross-traversal largely consistent static geometry and appearance that varies drastically with environmental conditions within a unified framework. This section first presents the formal definition of the task, followed by an overview of our proposed Appearance Decomposition Gaussian Splatting (ADM-GS) hybrid scene graph framework.

\begin{figure*}[t]
    \centering
    \includegraphics[width=1.0\linewidth]{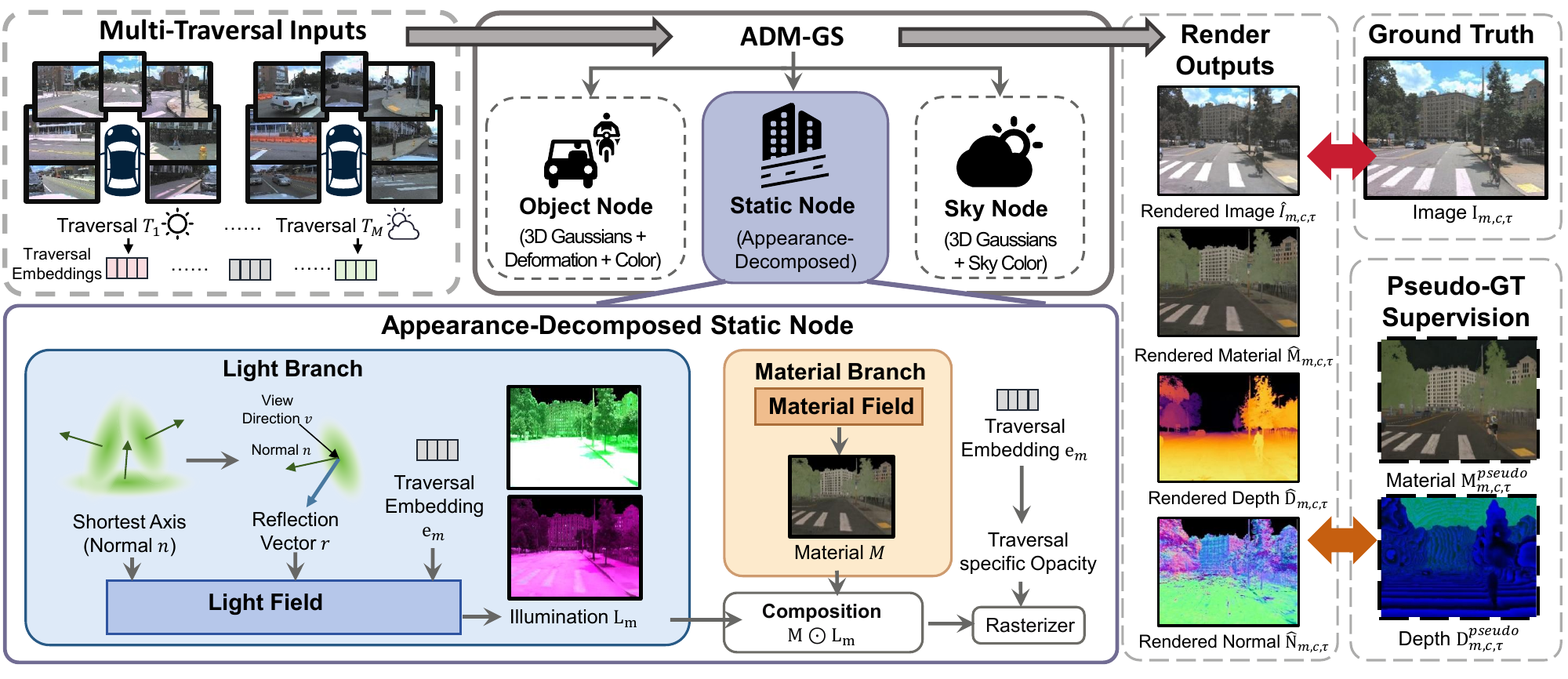} 

    \caption{\textbf{Framework of Appearance Decomposition Gaussian Splatting for Multi-Traversal Reconstruction.}
    ADM-GS represents the scene with a hybrid scene graph consisting of static, object, and sky nodes. Its core static node decomposes appearance into a traversal-invariant material field and a traversal-dependent light field, where normal and reflection-vector cues are introduced to improve illumination prediction. A traversal-conditioned opacity gating module, adapted from 4DGF~\cite{fischer2024dynamic}, uses the traversal embedding $e_m$ to predict traversal-specific opacity, which suppresses geometrically inconsistent regions during training. Ground-truth RGB images and pseudo supervision on material and normals are used for optimization.}
    
    \label{fig:method_overview}
\end{figure*}

\subsection{Problem Formulation}
\label{subsec:prblem_formulation}
    We define the input data as a collection of traversals $\mathcal{T}=\{T_1,T_2,\ldots,T_M\}$ captured from the same geographic area at different time periods. Each traversal $T_m$ consists of a set of observations recorded by a vehicle-mounted camera set $\mathcal{C}$. For an image $I_{m,c,\tau}$ captured in traversal $T_m$ by camera $c\in\mathcal{C}$ at timestamp $\tau$, the corresponding camera intrinsics $K_c$ and camera pose in the world coordinate system $P_{m,c,\tau}\in SE(3)$ are known. In addition, the scene contains a set of dynamic objects, each annotated with 3D bounding box trajectories over time. 
    
    Unlike single-sequence reconstruction, the key characteristic of multi-traversal data is that the static geometry is largely shared across traversals, while the appearance may vary due to illumination changes and sensor characteristics. Our goal is to learn a parameterized scene representation for multi-traversal reconstruction. Given a camera pose $P_{m,c,\tau}$, a timestamp $\tau$, and a traversal condition indexed by $m$, the representation synthesizes a target image $\hat{I}_{m,c,\tau}$ under the corresponding viewing and traversal conditions. To achieve this, the scene is modeled as a compositional rendering of the static background, dynamic objects, and the sky background.
    
\subsection{Framework Overview}
\label{subsec:framework_overview}

To effectively model the complex composition of multi-traversal autonomous driving scenes, we propose the Appearance Decomposition Gaussian Splatting (ADM-GS) framework. Building upon the efficient hybrid representation of 4DGF~\cite{fischer2024dynamic}, our framework leverages 3D Gaussians as an explicit geometry scaffold coupled with implicit neural fields for appearance modeling. Unlike prior works that learn an entangled radiance field, our framework reformulates the network architecture to enable explicit appearance decomposition. Specifically, our framework decomposes the scene into three types of nodes based on their attributes:
    
    \subsubsection{Appearance-Decomposed Static Node}
    This node represents the static background of the scene, such as road surfaces and building facades. To handle both structural changes and appearance variation within a unified framework, we design a two-layer architecture containing geometric gating and appearance decomposition. For transient geometry, we adopt the geometric encoding mechanism from 4DGF~\cite{fischer2024dynamic} to reduce the influence of geometric inconsistencies across traversals, such as temporary billboards. 
    This module predicts an opacity attenuation term via an independent MLP, which suppresses invalid regions when temporary objects appear or local material changes occur. On top of this geometric gating, we perform appearance decomposition by constructing two independent neural fields: a material field that takes spatial geometric features as input to predict traversal-invariant material, and a light field conditioned on normals, reflection vectors, and environmental codes to capture traversal-dependent shadows, specular highlights, and illumination variations.
    
    \subsubsection{Sky Node}
    As a background at infinity, the sky has no explicit 3D geometric surface, and its texture varies drastically with environmental changes. For this component, we do not employ appearance decomposition but model it as a neural field-based appearance module. Specifically, we place Gaussian primitives on a far-field sphere outside the scene boundaries. Their colors are directly predicted by a neural field taking the view direction and a traversal-specific appearance embedding as inputs. This approach flexibly models highly varying sky textures without requiring explicit appearance decomposition.
    
    \subsubsection{Object Node}
    For dynamic entities such as vehicles and pedestrians in the scene, we adopt object-centric modeling. Each dynamic object is instantiated as an independent node with its own coordinate system. To handle non-rigid motions, such as changes in pedestrian pose, we introduce a deformation module in the canonical space, which predicts Gaussian coordinate offsets based on timestamps, thereby mapping canonical geometry to the state at the observation time.
    
    \subsubsection{Rendering and Sensor Alignment}
    During rendering, we transform the 3D Gaussians of all nodes into the world coordinate system and render a blended scene image via an efficient tile-based rasterizer. To compensate for photometric discrepancies caused by auto-exposure and white balance across traversals, we further apply a learnable per-traversal affine transformation to the rasterized image:
    \begin{equation}
    \hat{\mathbf{I}}_{m,c,\tau}
    =
    \mathbf{s}_{m}\odot \mathbf{I}^{\mathrm{render}}_{m,c,\tau}
    +
    \mathbf{b}_{m},
    \end{equation}
    where $\mathbf{I}^{\mathrm{render}}_{m,c,\tau}$ denotes the rasterized image for camera $c$ at timestamp $\tau$ in traversal $T_m$, and $\mathbf{s}_{m}\in\mathbb{R}^{3}$ and $\mathbf{b}_{m}\in\mathbb{R}^{3}$ are traversal-specific scale and bias parameters. This lightweight alignment maps the rendered output to the image space of traversal $T_m$ and improves consistency between synthesized images and real observations.

\subsection{Appearance Decomposition for the Static Background}
\label{subsec:intrinsic_appearance}
In multi-traversal autonomous driving scenarios, the core challenge in reconstructing the static background is that its observed appearance fluctuates significantly due to varying illumination conditions, despite the material properties remaining constant. To address this, we adopt an appearance decomposition strategy for the static background. Our goal is to separate the observations into a traversal-dependent illumination representation that captures changing environmental effects, and a traversal-invariant material field that models stable material appearance across traversals.

    \subsubsection{Appearance Decomposition Formulation} 
    Our formulation is motivated by the Rendering Equation~\cite{kajiya1986rendering}, which provides a useful perspective for understanding that observed appearance contains both diffuse and view-dependent components. For a surface point $\mathbf{x}\in\mathbb{R}^3$ viewed from direction $\mathbf{v}\in\mathbb{S}^2$, the outgoing radiance can be written as
    \begin{equation}
    \label{eq:rendering_equation}
    \mathcal{L}_o(\mathbf{x}, \mathbf{v})
    =
    \int_{\Omega^+}
    \mathcal{L}_i(\mathbf{x}, \boldsymbol{\omega}_i)\,
    f_r(\mathbf{x}, \boldsymbol{\omega}_i, \mathbf{v})\,
    (\mathbf{n}\cdot \boldsymbol{\omega}_i)\,
    d\boldsymbol{\omega}_i,
    \end{equation}
    where $\Omega^+$ denotes the hemisphere oriented around the surface normal vector $\mathbf{n}$, $\boldsymbol{\omega}_i$ denotes the incident light direction, $\mathbf{v}$ denotes the viewing direction, and $\mathcal{L}_i(\mathbf{x}, \boldsymbol{\omega}_i)$ is the incident radiance arriving from direction $\boldsymbol{\omega}_i$. To motivate appearance decomposition, we write the bidirectional reflectance distribution function (BRDF) as the sum of a diffuse term and a specular term:
    \begin{equation}
    \label{eq:brdf}
    f_r(\mathbf{x}, \boldsymbol{\omega}_i, \mathbf{v})
    =
    f_d(\mathbf{x}, \boldsymbol{\omega}_i)
    +
    f_s(\mathbf{x}, \boldsymbol{\omega}_i, \mathbf{v}),
    \end{equation}
    where $f_d$ denotes the diffuse component and $f_s$ denotes the specular component.
    
    In large-scale multi-traversal driving scenes, explicitly solving Eq.~(\ref{eq:rendering_equation}) is intractable due to the absence of precise lighting priors and the complexity of outdoor illumination. Therefore, rather than performing full inverse rendering, we adopt a local factorized approximation inspired by appearance decomposition. Specifically, we factor out a traversal-invariant material-like term and absorb the remaining directional transport effects into an aggregated illumination term:
    \begin{equation}
    \label{eq:illumination_bridge}
    \mathbf{L}(\mathbf{x}, \mathbf{v})
    =
    \int_{\Omega^+}
    \mathcal{L}_i(\mathbf{x}, \boldsymbol{\omega}_i)\,
    \tilde f(\mathbf{x}, \boldsymbol{\omega}_i, \mathbf{v})\,
    (\mathbf{n}\cdot \boldsymbol{\omega}_i)\,
    d\boldsymbol{\omega}_i,
    \end{equation}
    where $\tilde f$ denotes an effective directional response that summarizes the interaction of the diffuse and specular components in Eq.~(\ref{eq:brdf}). Under this approximation, the rendered image appearance can be written as
    \begin{equation}
    \label{eq:factorized_color}
    \hat{\mathbf{I}}(\mathbf{x}, \mathbf{v}, m)
    \approx
    \mathbf{M}(\mathbf{x}) \odot \mathbf{L}(\mathbf{x}, \mathbf{v}).
    \end{equation}
    In our model, $\mathbf{M}(\mathbf{x})$ is instantiated as a traversal-invariant material field, while $\mathbf{L}$ is represented by a unified neural light field that absorbs traversal-dependent illumination effects, including both diffuse illumination variation and view-dependent highlights. The final rendered image appearance is formulated as
    \begin{equation}
    \hat{\mathbf{I}}(\mathbf{x}, \mathbf{v}, m)
    = 
    \mathbf{M}(\mathbf{x}) \odot \mathbf{L}(\mathbf{x}, \mathbf{n}, \mathbf{v}, e_m),
    \end{equation}
    where $\odot$ denotes the Hadamard product.

    \subsubsection{Decoupled Static Material Field}
    To encourage a consistent decomposition, the material prediction is architecturally decoupled from any view-dependent or traversal-dependent cues. We define the material mapping $\Phi_{\mathrm{material}}$ as:
    
    \begin{equation}
    \mathbf{M}(\mathbf{x}) = \Phi_{\mathrm{material}}(f_{\mathrm{geo}}(\mathbf{x}))
    \quad \text{s.t.} \quad
    \mathbf{M}(\mathbf{x}) \in [0,1]^3,
    \end{equation}
    where $f_{\mathrm{geo}}(\mathbf{x})$ denotes spatial geometric features derived from the Gaussian field.

    Optimizing this static material field on dynamic sequences requires handling geometric inconsistencies such as transient obstacles. To this end, we leverage the transient geometry mechanism~\cite{fischer2024dynamic}, which modulates the opacity of the underlying 3D Gaussians. Since opacity determines the visibility of a Gaussian in the final rendering, suppressing the opacity of geometrically inconsistent regions effectively excludes them from the compositing process. This mechanism helps restrict material optimization to geometrically consistent observations, reducing the influence of transient objects or local background changes.
    
    \subsubsection{Neural Light Field}
    The illumination term $\mathbf{L}$ captures the interaction between environmental lighting and surface geometry. We define the illumination mapping $\Phi_{\mathrm{light}}$ as:
    
    \begin{equation}
    \begin{aligned}
    \mathbf{L}(\mathbf{x}, \mathbf{n}, \mathbf{v}, e_m)
    &= \Phi_{\mathrm{light}}(\mathbf{n}, \mathbf{v}, f_{\mathrm{geo}}(\mathbf{x}), e_m) \\
    &\quad \text{s.t.} \quad \mathbf{L}(\mathbf{x}, \mathbf{n}, \mathbf{v}, e_m)\in\mathbb{R}_+^3,
    \end{aligned}
    \end{equation}
    
    where the function explicitly depends on the surface normal $\mathbf{n}$ and view direction $\mathbf{v}$ to model view-dependent lighting effects. To capture the varying environmental illumination across different traversals, we condition the network on a learnable traversal-specific embedding $e_m$. The output is constrained to the positive domain, allowing the light field to represent a wide dynamic range of illumination effects. Specific parameterization strategies to effectively capture high-frequency reflections are detailed in Section~\ref{subsec:geo_reg}.

\subsection{Geometric Regularization and Reflection Modeling}
\label{subsec:geo_reg}
    
    The effectiveness of the appearance decomposition introduced in Section~\ref{subsec:intrinsic_appearance} hinges on two critical factors: stable surface normals for diffuse calculation and the capability to model view-dependent highlights. However, the original 3D Gaussian Splatting lacks explicit geometric constraints, often resulting in loosely structured volumetric geometry with noisy normals that degrade illumination estimation. Instead of relying solely on raw view direction inputs, we additionally introduce reflection direction reparameterization to better capture high-frequency reflection dynamics. Accordingly, we combine a geometric regularization term for surface-aware geometry with reflection-aware reparameterization for illumination prediction.
    
    \subsubsection{Surface-Aware Geometric Regularization}
        To obtain stable normals $n$ from discrete 3D Gaussian primitives, the Gaussian geometry should approximate surfels aligned with local surfaces, rather than volumetric blobs that fill the surrounding space.

        We introduce a scale-based flatness constraint. For each Gaussian primitive $k$, its covariance matrix is parameterized by rotation $q_k$ and scaling $s_k\in\mathbb{R}^{3}$. To encourage each Gaussian to compress along its shortest axis, we impose an anisotropic penalty on the scaling parameters in logarithmic space:
        
        \begin{equation}
        \mathcal{L}_{\text{scale}}=
        \frac{1}{|\mathcal{G}|}\sum_{k\in\mathcal{G}}
        \max\!\left(0,\delta-\big(\log(s_{k}^{\max})-\log(s_{k}^{\min})\big)\right),
        \end{equation}
        
        where $s_{k}^{\min}$ and $s_{k}^{\max}$ are the minimum and maximum values in the scaling vector $s_k$, respectively, and $\delta$ is the flatness threshold. In our implementation, we set $\delta = 1.0$, which corresponds to penalizing Gaussians with $s_{k}^{\min}/s_{k}^{\max} > e^{-1} \approx 0.37$, i.e., Gaussians that are not sufficiently flat. By penalizing Gaussian primitives with insufficient anisotropy, this loss encourages them to become flatter ellipsoids aligned with local surfaces.
        
        The shortest-axis direction $u_k^{\min}$ is used to estimate the geometric normal $n_k$ for each Gaussian primitive. We directly use this geometric normal in the illumination computation during rendering, avoiding the need for an additional normal prediction network. During rasterization, these per-Gaussian normals are composited over the static background to form a rendered background normal map, denoted as $N_{\text{pred}}$, which is later used for supervision.
    
    \subsubsection{Reflection Direction Reparameterization}
        Standard neural representations typically condition color prediction directly on the viewing direction $\mathbf{v}$. However, as analyzed in Ref-NeRF~\cite{verbin2024ref}, this requires the MLP to implicitly learn the complex angular transformation between the viewing direction and the local reflection frame, leading to difficulties in interpolating high-frequency specular highlights.
        
        Motivated by reflection-aware view-dependent appearance modeling, we adapt reflection direction reparameterization to our light field for multi-traversal scenes. Specifically, we compute the reflection vector $\mathbf{r}$ about the surface normal $\mathbf{n}$:
        \begin{equation}
        \mathbf{r} = 2 (\mathbf{n} \cdot \mathbf{v}) \mathbf{n} - \mathbf{v}.
        \end{equation}
        By feeding $\mathbf{r}$ into the light field, we align the input domain with the peak direction of the specular lobe. This effectively simplifies the network's task from learning a spatially-varying rotation function to approximating a smoother mapping of the environment lighting, helping the network model sharper view-dependent highlights.

\begin{figure}[t] 
    \centering
    \includegraphics[width=0.85\linewidth]{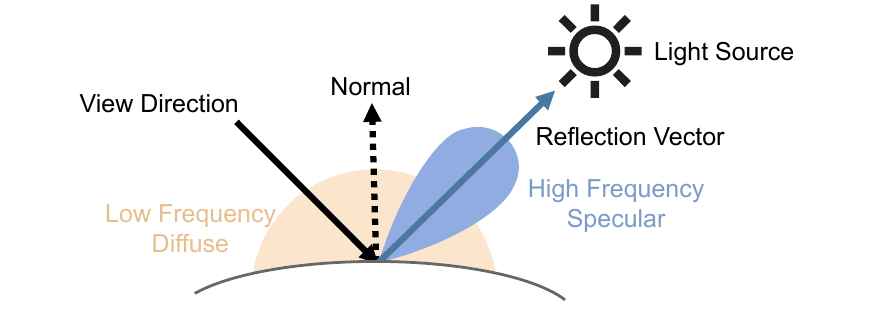} 
    
    \caption{\textbf{Illustration of frequency-separated illumination cues.}
    Surface normals are mainly associated with low-frequency diffuse illumination, while reflection vectors are more informative for high-frequency view-dependent specular effects.}
    \label{fig:reflection_vector}
\end{figure}

    \subsubsection{Frequency-Separated Hybrid Encoding}
        Diffuse and specular reflections exhibit significantly different characteristics in the frequency domain: diffuse reflection is primarily influenced by low-frequency surface orientation, whereas specular reflection contains extremely high-frequency view-dependent details, as illustrated in Fig.~\ref{fig:reflection_vector}. To match this property, we design a frequency-separated feature encoding strategy to drive the illumination MLP:
        
        \vspace{0.5em}
        \indent \textit{Low-Frequency Normal Encoding:} We encode the normal $n$ using only 1st-order spherical harmonics to represent the diffuse component in Eq.~\ref{eq:brdf}. Diffuse illumination $I_d \propto \max(0,n\cdot l)$ is a low-frequency function of the normal, where $l$ denotes the light direction. Limiting the encoding frequency is sufficient to capture the geometric dependence of diffuse reflection while effectively filtering out high-frequency normal noise residual from gaussian geometric reconstruction, generating a smooth diffuse base.
        
        \indent \textit{High-Frequency Reflection Encoding:} For the reflection vector $r$, we employ 4th-order high-frequency spherical harmonic encoding to represent the specular component in Eq.~\ref{eq:brdf}. This endows the network with extremely high angular resolution, enabling it to acutely capture sharp highlights distributed along the reflection direction and accurately restore the texture of materials such as metal and glass.

        \indent \textit{Fusion Mechanism:}
        The final illumination feature space is formed by concatenating the aforementioned encodings with spatial context features, which is then fed into $MLP_{\mathrm{light}}$:
        \begin{equation}
        \begin{aligned}
        f_{\mathrm{in}}(\mathbf{x},\mathbf{v},m)
        &=
        \operatorname{Concat}\!\left[
        \gamma_{\mathrm{deg}=1}(\mathbf{n}),
        \gamma_{\mathrm{deg}=4}(\mathbf{r}),
        \gamma_{\mathrm{deg}=3}(\mathbf{v}),\right.\\
        &\qquad\left.
        f_{\mathrm{geo}}(\mathbf{x}),
        e_m
        \right].
        \end{aligned}
        \end{equation}
        The fused feature is used to predict the illumination/light field $\mathbf{L}(\mathbf{x},\mathbf{n},\mathbf{v},\mathbf{r},e_m)$ through $MLP_{\mathrm{light}}$, where $\gamma^{\mathrm{deg}=k}(\cdot)$ denotes the spherical harmonics (SH) encoding truncated at degree $k$. Specifically, $\gamma^{\mathrm{deg}=1}(\mathbf{n})$ provides a low-frequency encoding of the surface normal for modeling the diffuse component, while $\gamma^{\mathrm{deg}=4}(\mathbf{r})$ provides a higher-frequency angular encoding of the reflection direction for capturing view-dependent specular effects. $\gamma^{\mathrm{deg}=3}(\mathbf{v})$ denotes the degree-3 encoding of the viewing direction $\mathbf{v}$, $f_{\mathrm{geo}}(\mathbf{x})$ denotes the spatial geometric feature at 3D point $\mathbf{x}$, and $e_m$ denotes the traversal embedding associated with traversal $T_m$.

\subsection{Optimization and Supervision}
\label{subsec:optimization}

To jointly optimize the geometric structure, dynamic neural fields, and appearance-related attributes of static nodes in the scene graph, we employ a unified differentiable rendering framework. Our optimization objective encourages the synthesized images to remain photometric consistency with the observations, while regularizing the decomposition through geometric and pseudo-supervised priors.

The total loss function consists of a photometric reconstruction loss, decomposition supervision losses, and a geometric regularization term:
\begin{equation}
\begin{aligned}
\mathcal{L}
&=\mathcal{L}_{\mathrm{photo}}+\lambda_{\mathrm{decomp}}\left(\mathcal{L}_{\mathrm{material}}+\mathcal{L}_{\mathrm{normal}}\right) \\
&\quad +\lambda_{\mathrm{scale}}\mathcal{L}_{\mathrm{scale}}.
\end{aligned}
\end{equation}

We combine L1 loss and SSIM loss to supervise the consistency between the final rendered image $\hat{\mathbf{I}}_{m,c,\tau}$ and the observed image $\mathbf{I}_{m,c,\tau}$:
\begin{equation}
\begin{aligned}
\mathcal{L}_{\mathrm{photo}}
&=(1-\lambda_{\mathrm{ssim}})\left\|\hat{\mathbf{I}}_{m,c,\tau}-\mathbf{I}_{m,c,\tau}\right\|_1 \\
&\quad +\lambda_{\mathrm{ssim}}\left(1-\mathrm{SSIM}\left(\hat{\mathbf{I}}_{m,c,\tau},\mathbf{I}_{m,c,\tau}\right)\right)
\end{aligned}
\end{equation}

To regularize the ill-posed material--illumination decomposition, we introduce pseudo supervision based on monocular estimation. Stable surface normals provide a geometric basis for illumination estimation. Since raw LiDAR point clouds are sparse and do not directly provide dense normal supervision, we use UniDepthV2~\cite{piccinelli2025unidepthv2} to generate monocular depth maps $\mathbf{D}^{\mathrm{pseudo}}_{m,c,\tau}$ and derive pseudo normal maps $\mathbf{N}^{\mathrm{pseudo}}_{m,c,\tau}$ from their spatial gradients. We impose a masked L1 constraint on the rendered background normal map $\hat{\mathbf{N}}_{m,c,\tau}$, which is obtained by rasterizing the geometric normals defined in Section~\ref{subsec:geo_reg} over the visible static background:
\begin{equation}
\mathcal{L}_{\mathrm{normal}}=\left\|\left(\hat{\mathbf{N}}_{m,c,\tau}-\mathbf{N}^{\mathrm{pseudo}}_{m,c,\tau}\right)\odot\mathrm{mask}^{\mathrm{normal}}_{m,c,\tau}\right\|_1
\end{equation}
where $\mathrm{mask}^{\mathrm{normal}}_{m,c,\tau}$ denotes the supervision mask over static-background pixels with valid pseudo depth.

For material supervision, we use material maps generated by Ouroboros~\cite{sun2025ouroboros} as pseudo supervision $\mathbf{M}^{\mathrm{pseudo}}_{m,c,\tau}$. We impose supervision on the rendered material map $\hat{\mathbf{M}}_{m,c,\tau}$ of the static material field only in valid static-background regions:
\begin{equation}
\mathcal{L}_{\mathrm{material}}=\left\|\left(\hat{\mathbf{M}}_{m,c,\tau}-\mathbf{M}^{\mathrm{pseudo}}_{m,c,\tau}\right)\odot\mathrm{mask}^{\mathrm{material}}_{m,c,\tau}\right\|_1
\end{equation}
where $\mathrm{mask}^{\mathrm{material}}_{m,c,\tau}$ denotes the supervision mask over static-background pixels with available pseudo material values.

As described in Section~\ref{subsec:geo_reg}, we introduce the flatness constraint $\mathcal{L}_{scale}$ to encourage 3D Gaussian primitives to become flatter, surface-aligned ellipsoids. This not only stabilizes the estimation of explicit normals, but also provides a more suitable geometric basis for the normal supervision described above.

\section{EXPERIMENTS}

\subsection{Experimental Settings}
\label{sec:settings}

    \subsubsection{Datasets}
    We evaluate our method on two large-scale autonomous driving datasets: Argoverse 2~\cite{wilson2023argoverse} and the Waymo Open Dataset~\cite{sun2020scalability}. Argoverse 2 features sensor-rich scenes collected in six U.S. cities. For single-traversal evaluations, we follow the standard validation split used in recent benchmarks~\cite{wang2025unifying}. For multi-traversal evaluations, we select sequences in which the ego-vehicle revisits the same location at different times of day, resulting in noticeable illumination changes and shadow shifts. For the Waymo Open Dataset, we use the perception subset, which provides high-resolution imagery and LiDAR data. Following established protocols~\cite{yan2024street}, we evaluate our method on diverse driving scenarios to assess its performance across different sensor configurations and environments.
    
    \subsubsection{Baselines}
    To assess the performance of ADM-GS, we compare it against representative recent methods for autonomous driving scene reconstruction. For single-traversal settings, we compare with several recent 3DGS-based methods for driving scenes, including StreetGS~\cite{yan2024street}, OmniRe~\cite{chen2025omnire}, and Bilateral-Driving~\cite{wang2025unifying}. We also include PVG~\cite{chen2023periodic}, a dynamic 3DGS method that models scene dynamics with periodic vibration. For multi-traversal reconstruction, we primarily compare against 4DGF~\cite{fischer2024dynamic} and MTGS~\cite{li2025mtgs}. Both methods are designed to handle appearance changes across repeated traversals in large-scale driving environments, and therefore provide strong baselines for evaluating cross-traversal appearance modeling.
    
    \subsubsection{Implementation Details}
    We implement our method based on the 4DGF framework~\cite{fischer2024dynamic}, extending it with our appearance decomposition module and geometric regularization. All experiments are conducted on workstations equipped with four NVIDIA RTX 4090 GPUs. The training strategy is adjusted for different experimental settings.
    
    \noindent\textit{Single-traversal settings.}
    For standard scene reconstruction benchmarks, we adopt dataset-specific protocols. On Argoverse 2, following previous works~\cite{wang2025unifying, chen2025omnire}, we downsample the images by a factor of 3, resulting in a resolution of approximately $516 \times 682$, and train the model for 30,000 iterations. On the Waymo Open Dataset, we downsample the images by a factor of 2, resulting in a resolution of approximately $640 \times 960$, and train for 60,000 iterations to accommodate the higher resolution and scene complexity.
    
    \noindent\textit{Multi-traversal settings.}
    On Argoverse 2, we follow the multi-traversal training setup of 4DGF~\cite{fischer2024dynamic}, using the original full-resolution images without downsampling and training the model for 100,000 iterations.
    
\subsection{Comparison Results}


    \begin{table*}[t]
    \centering
    \caption{Quantitative comparison on single-traversal scenarios. We report results for novel view synthesis and scene reconstruction. Bold and underlined values indicate the best and second-best results, respectively.}
    
    \label{tab:single_travel_compare}
    \renewcommand{\arraystretch}{1.15} 
    \setlength{\tabcolsep}{8pt} 
    
    \begin{tabular}{ll ccc ccc}
    \toprule
    \multirow{2}{*}{\textbf{Dataset}} & \multirow{2}{*}{\textbf{Method}} & \multicolumn{3}{c}{\textbf{Novel View Synthesis}} & \multicolumn{3}{c}{\textbf{Scene Reconstruction}} \\
    \cmidrule(lr){3-5} \cmidrule(lr){6-8} 
     & & PSNR $\uparrow$ & SSIM $\uparrow$ & LPIPS $\downarrow$ & PSNR $\uparrow$ & SSIM $\uparrow$ & LPIPS $\downarrow$ \\
    \midrule
    \multirow{6}{*}{\textbf{Argoverse 2}} 
    & PVG~\cite{chen2023periodic} & 23.04 & 0.782 & 0.290 & 24.37 & 0.824 & 0.275 \\
    & StreetGS~\cite{yan2024street} & 23.40 & 0.795 & 0.197 & 25.02 & 0.857 & 0.181 \\
    & OmniRe~\cite{chen2025omnire} & 23.61 & 0.797 & 0.194 & 25.31 & 0.859 & 0.176 \\
    & Bilateral-Driving~\cite{wang2025unifying} & 23.70 & 0.794 & 0.183 & 25.74 & 0.865 & 0.164 \\
    & 4DGF \cite{fischer2024dynamic} & \underline{29.48} & \underline{0.890} & \underline{0.155} & \underline{30.17} & \underline{0.902} & \textbf{0.146} \\
    & \textbf{ADM-GS (Ours)} & \textbf{30.14} & \textbf{0.896} & \textbf{0.154} & \textbf{30.72} & \textbf{0.906} & \textbf{0.146} \\
    \midrule
    \multirow{6}{*}{\textbf{Waymo Open}} 
    & PVG~\cite{chen2023periodic} & 24.97 & 0.717 & 0.330 & 28.26 & 0.819 & 0.311 \\
    & StreetGS~\cite{yan2024street} & 24.48 & 0.707 & 0.276 & 27.67 & 0.824 & 0.255 \\
    & OmniRe~\cite{chen2025omnire} & 24.75 & 0.710 & 0.261 & 28.34 & 0.828 & 0.240 \\
    & Bilateral-Driving~\cite{wang2025unifying} & 24.91 & 0.713 & \underline{0.223} & \underline{29.29} & \underline{0.852} & \underline{0.195} \\
    & 4DGF \cite{fischer2024dynamic} & \underline{27.61} & \underline{0.848} & 0.225 & 28.83 & 0.866 & 0.216 \\
    & \textbf{ADM-GS (Ours)} & \textbf{28.91} & \textbf{0.862} & \textbf{0.186} & \textbf{29.93} & \textbf{0.883} & \textbf{0.176} \\
    \bottomrule
    \end{tabular}
    \end{table*}

    \begin{figure*}[t]
        \centering
        \setlength{\tabcolsep}{1pt}
        \renewcommand{\arraystretch}{0.5} 
        \newcolumntype{Y}{>{\centering\arraybackslash}m{0.185\linewidth}}
        \newcolumntype{Z}{>{\centering\arraybackslash}m{0.03\linewidth}}
        \begin{tabular}{Z Y Y Y Y Y}
            & {\small \textbf{GT}} & {\small \textbf{Ours}} & {\small \textbf{4DGF~\cite{fischer2024dynamic}}} & {\small \textbf{Bilateral-Driving~\cite{wang2025unifying}}} & {\small \textbf{OmniRe~\cite{chen2025omnire}}} \\
            \noalign{\vspace{3pt}} 
            \rotatebox{90}{\small \textbf{Argoverse 2}} &
            \includegraphics[width=\linewidth]{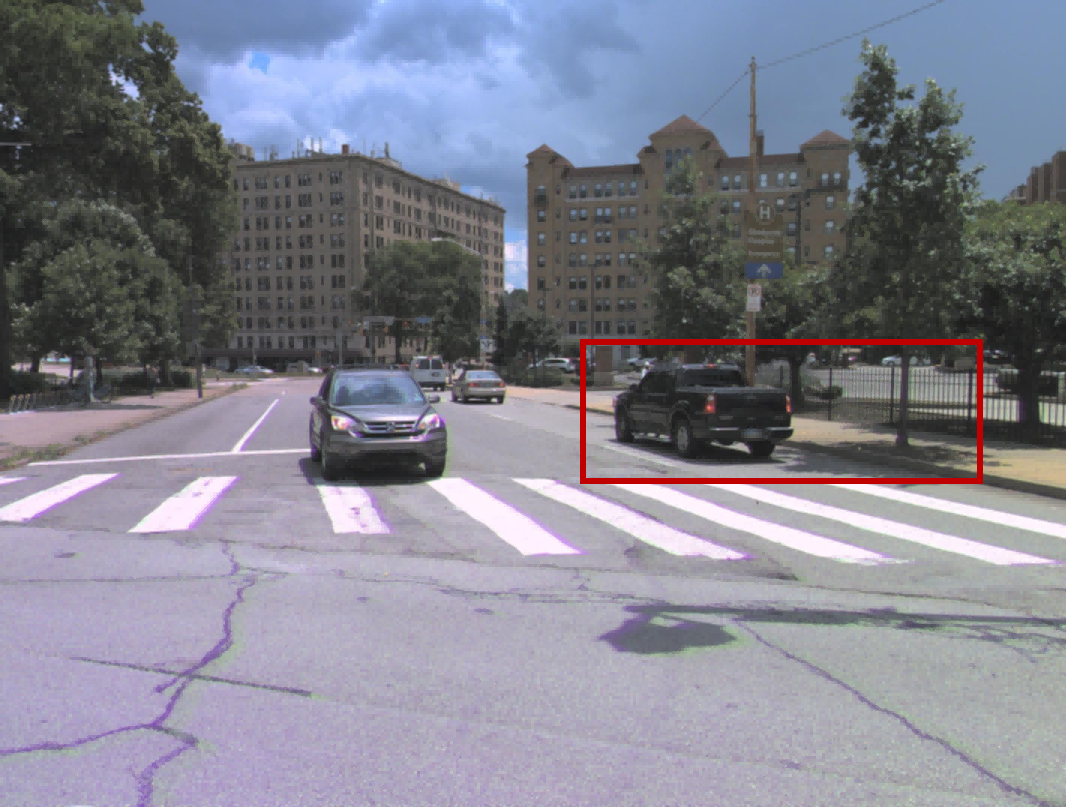} &
            \includegraphics[width=\linewidth]{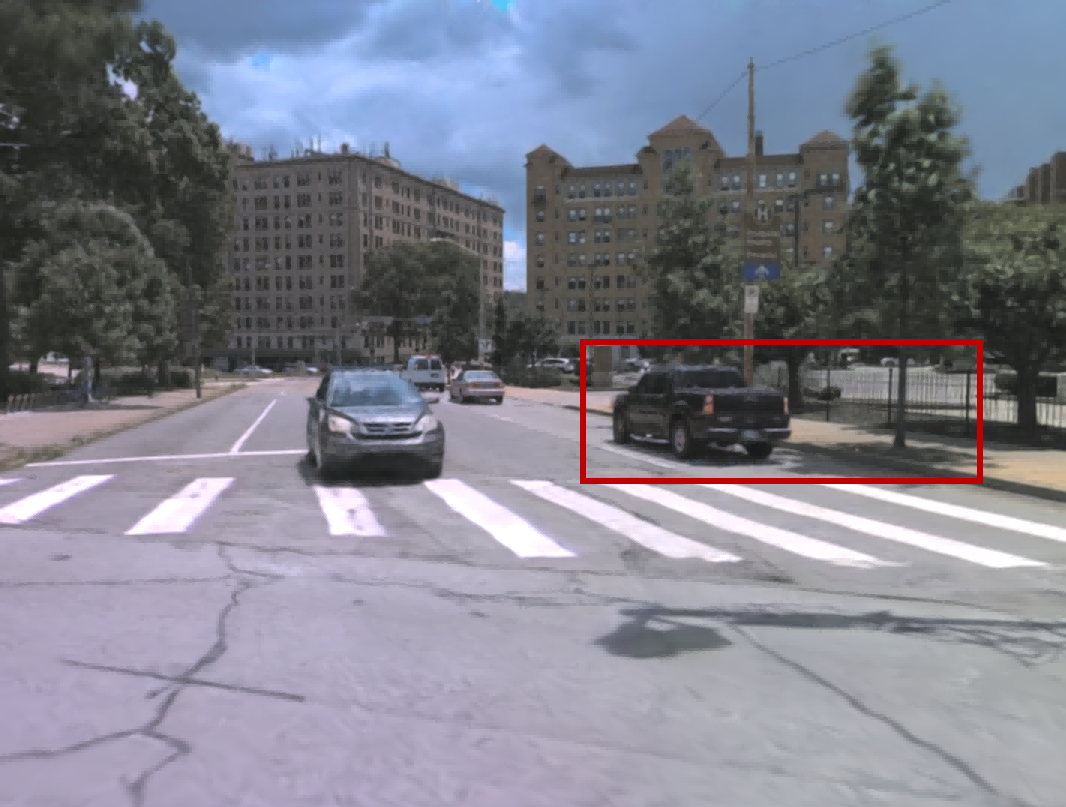} &
            \includegraphics[width=\linewidth]{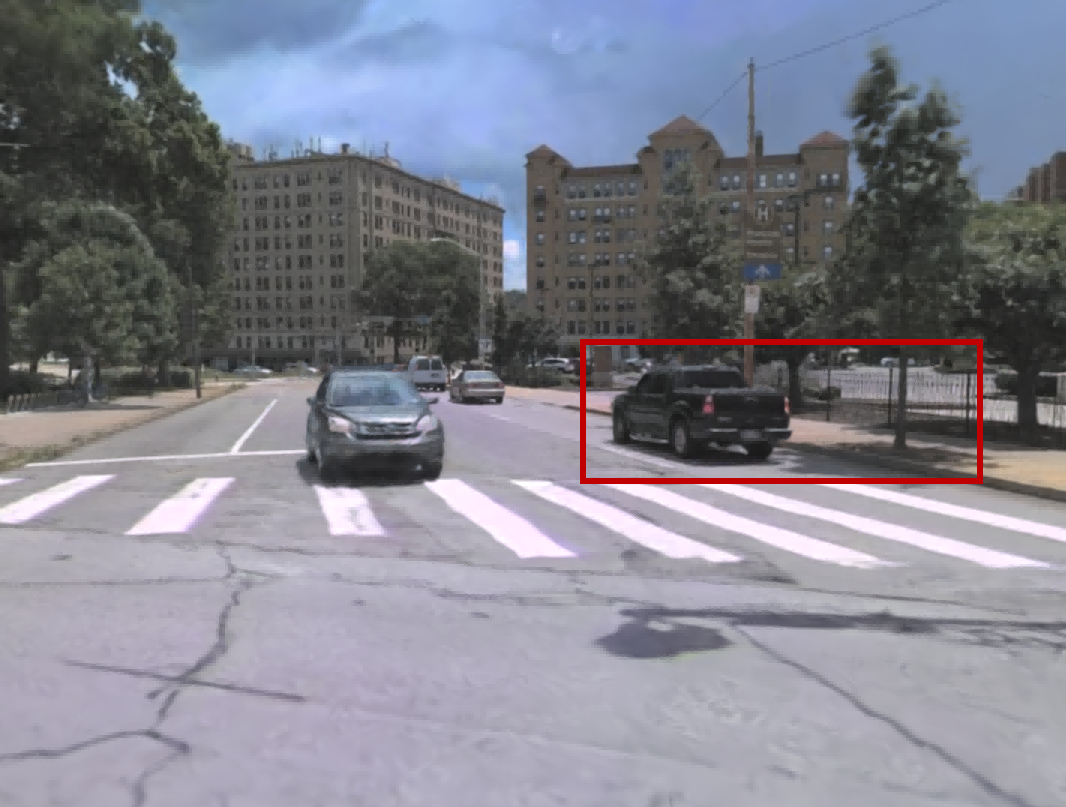} &
            \includegraphics[width=\linewidth]{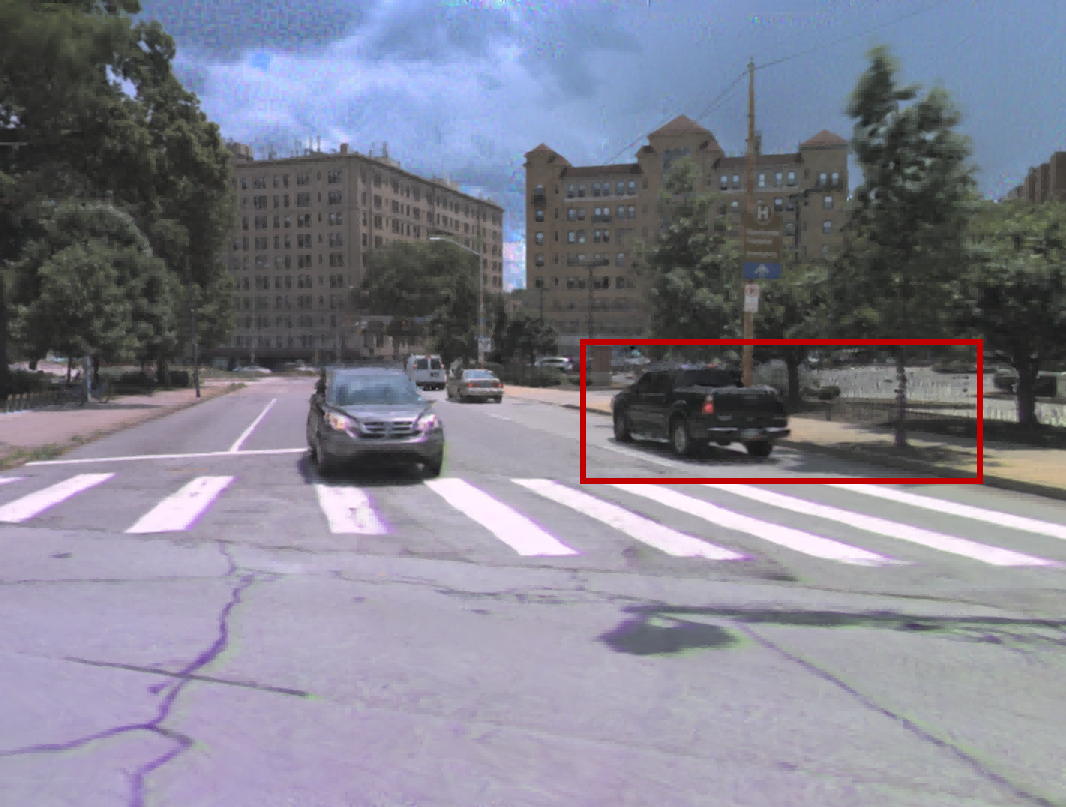} &
            \includegraphics[width=\linewidth]{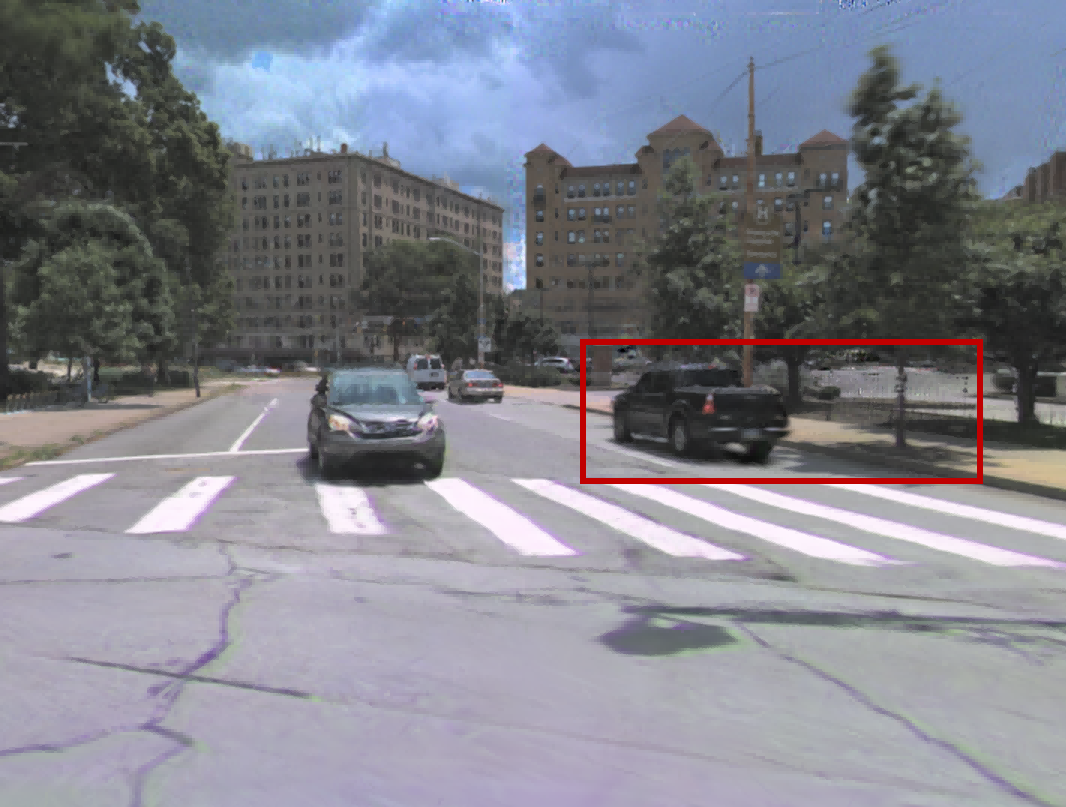} \\[-0.5ex] 
            \rotatebox{90}{\small \textbf{Waymo}} &
            \includegraphics[width=\linewidth]{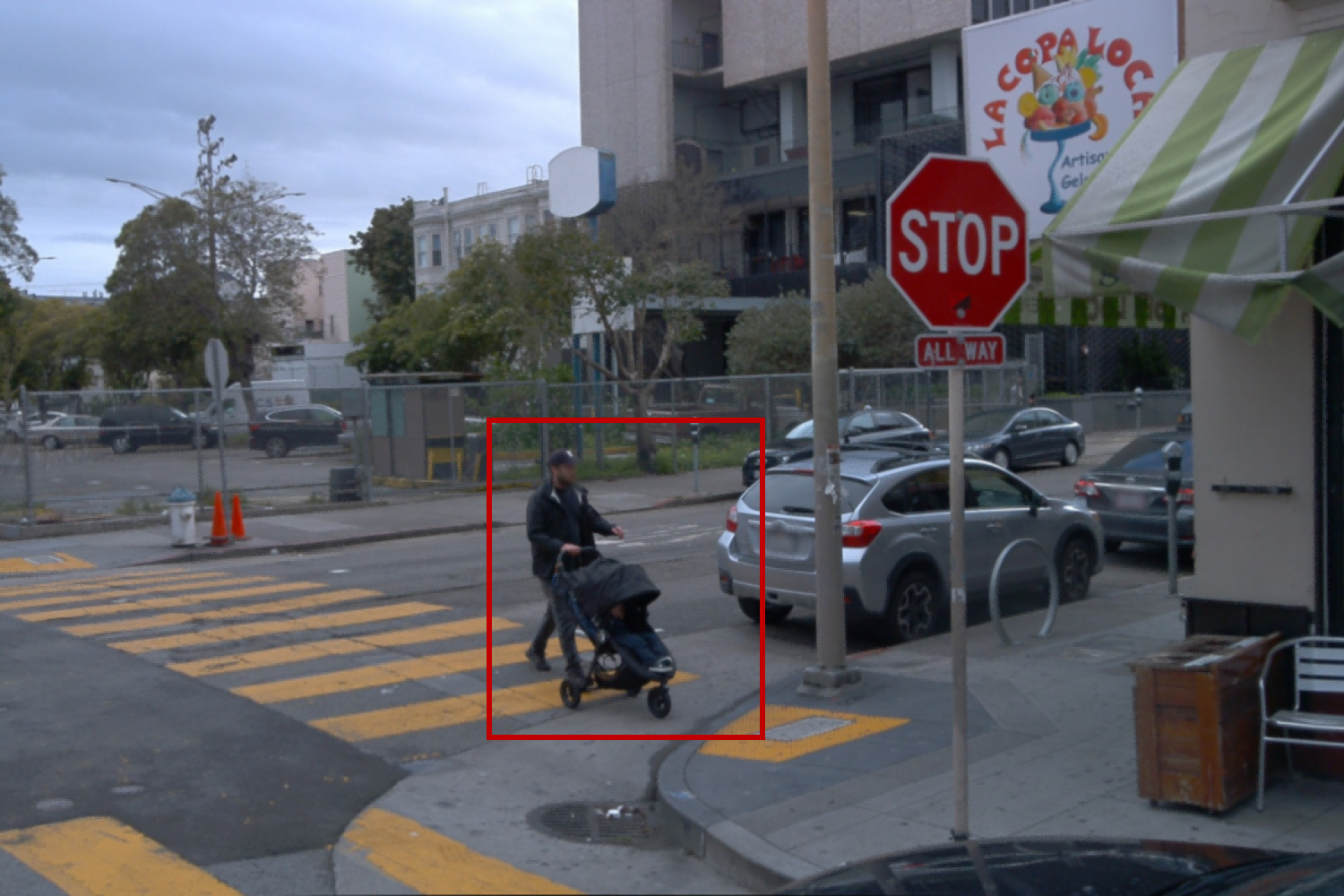} &
            \includegraphics[width=\linewidth]{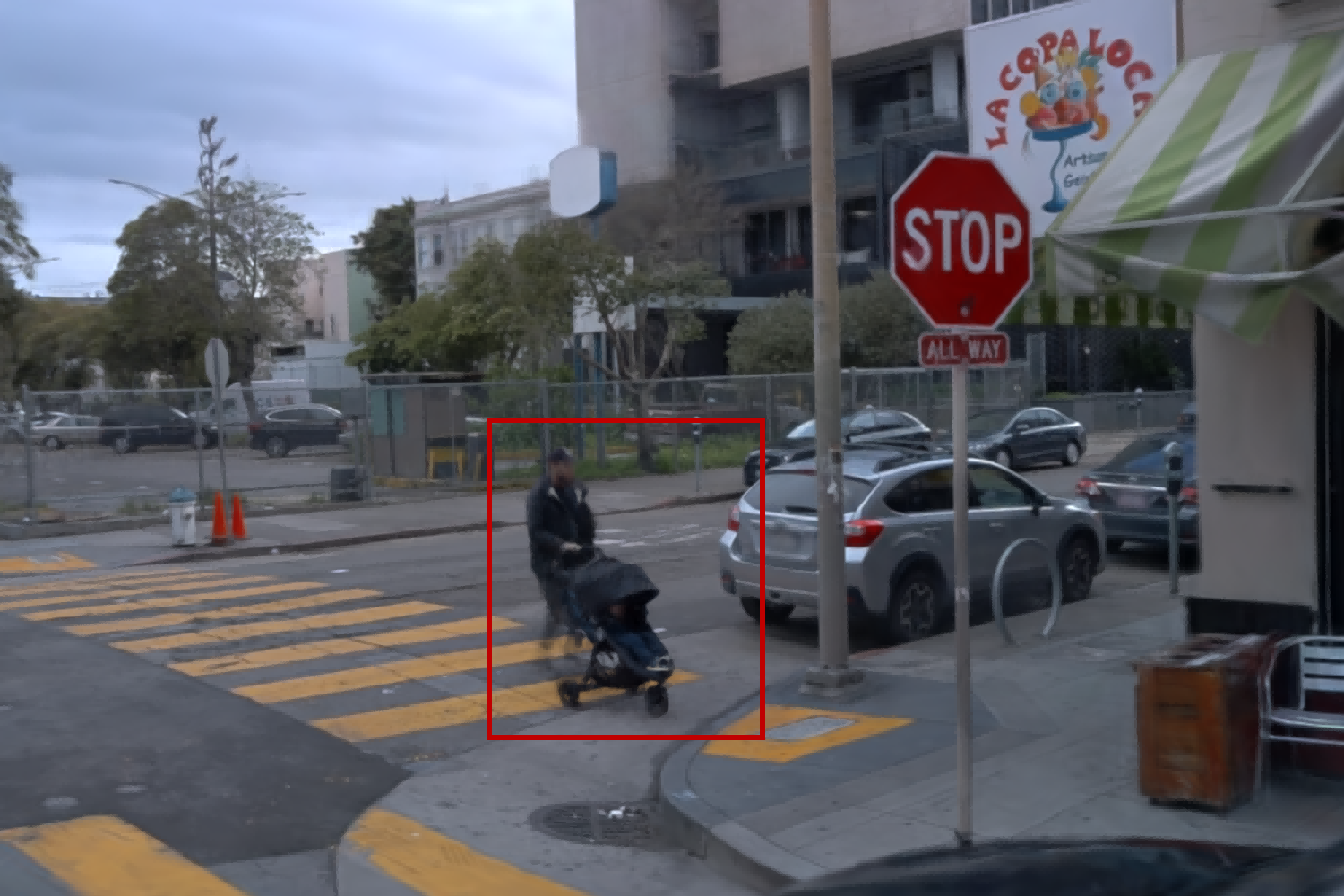} &
            \includegraphics[width=\linewidth]{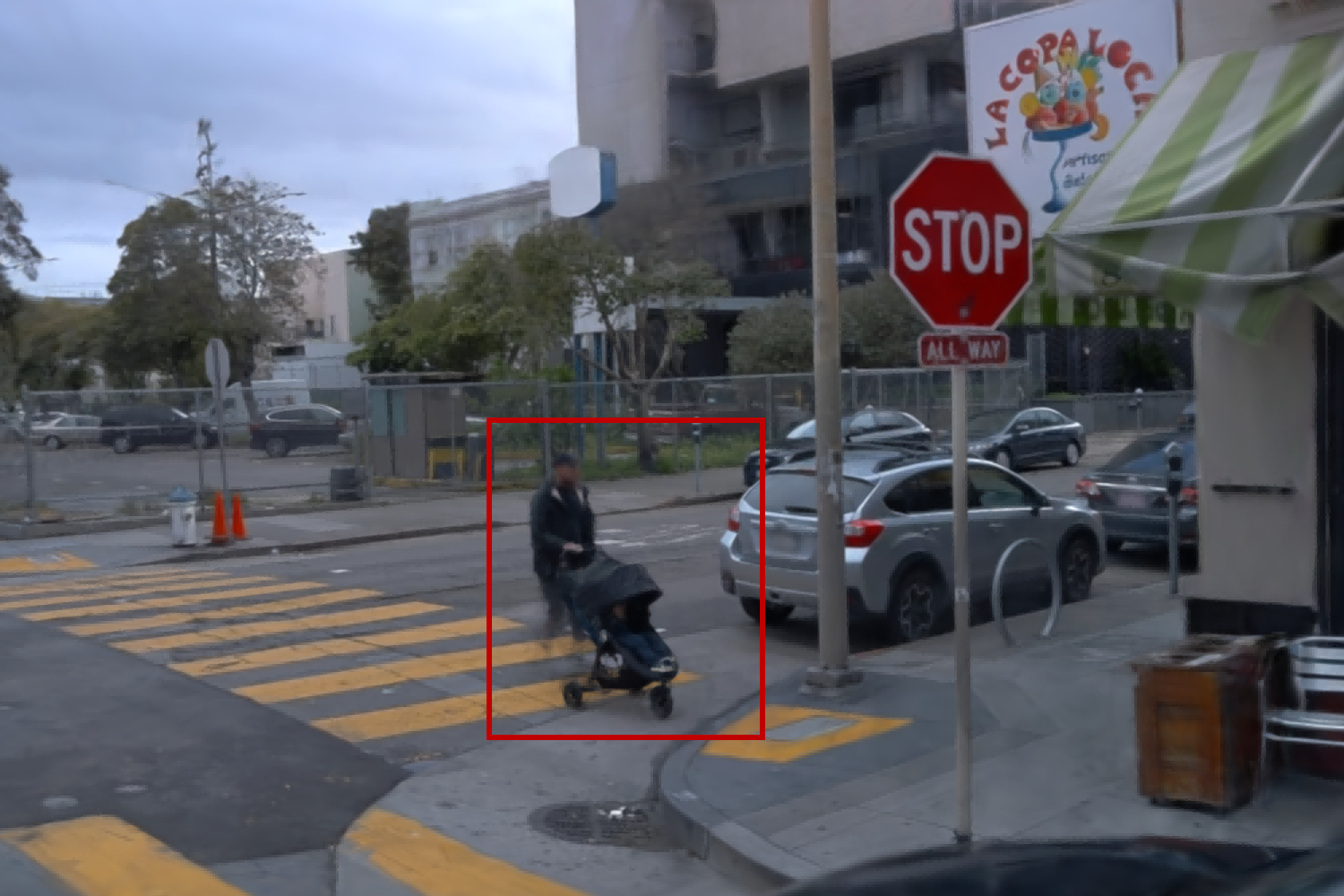} &
            \includegraphics[width=\linewidth]{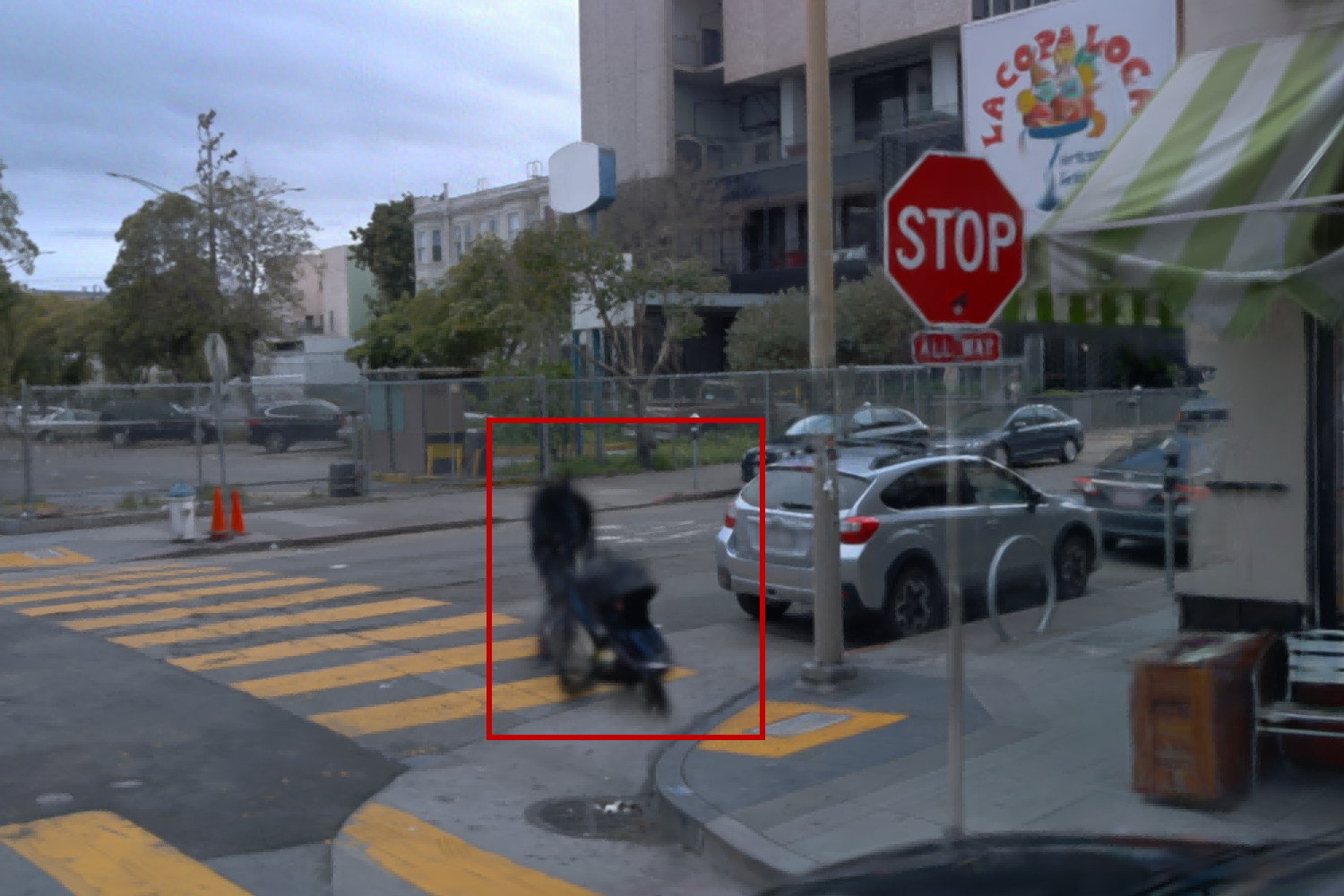} &
            \includegraphics[width=\linewidth]{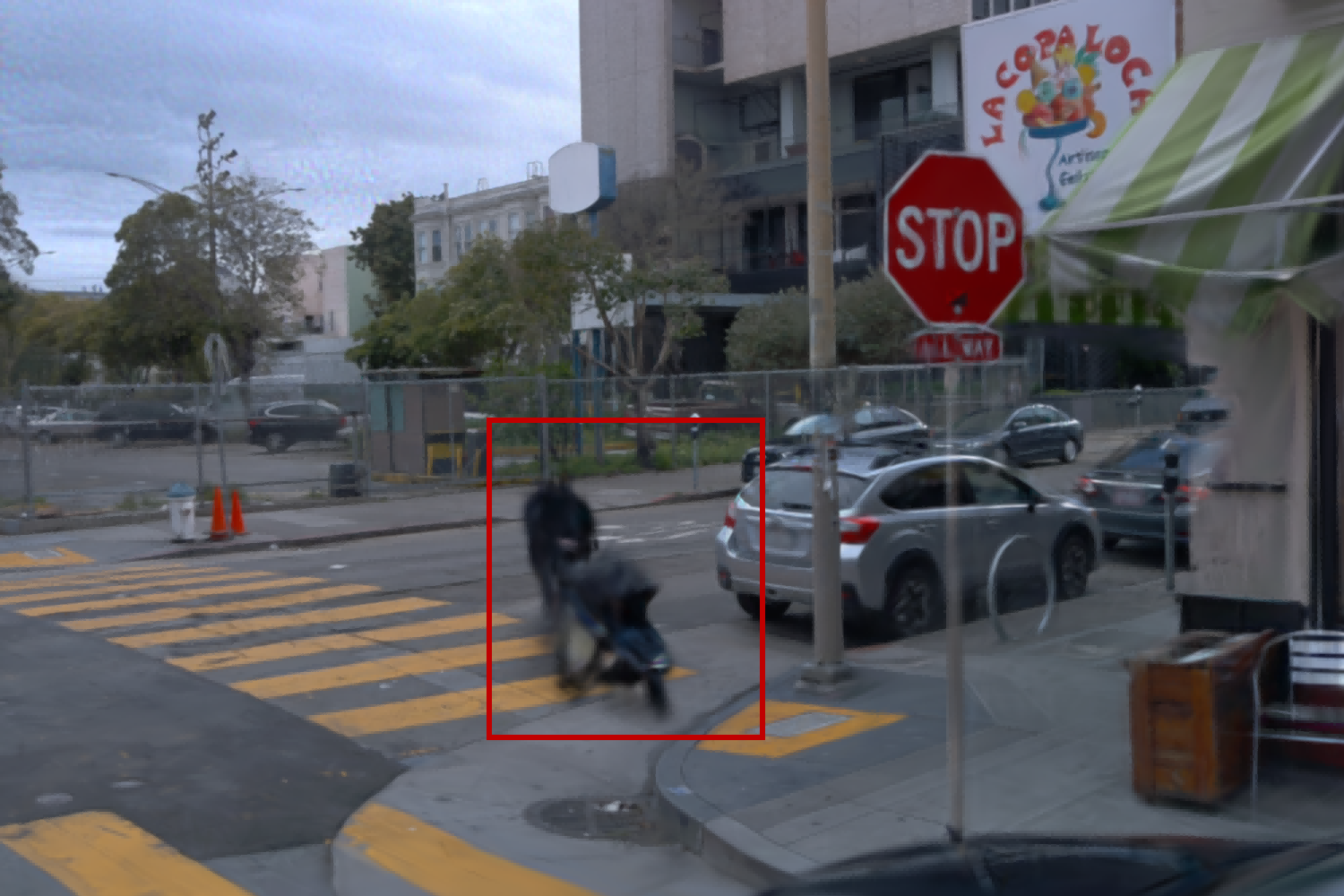} \\
        \end{tabular}
        \caption{\textbf{Visual comparison of novel view synthesis on dynamic scenes.} We compare our method with recent approaches on the Argoverse 2 (top row) and Waymo Open (bottom row) datasets. Methods such as Bilateral-Driving~\cite{wang2025unifying} and OmniRe~\cite{chen2025omnire} show noticeable blur and ghosting artifacts in dynamic-object regions. In contrast, our method produces sharper novel-view renderings and is visually closer to the ground truth.}
        \label{fig:vis_single_travel_nvs}
    \end{figure*}

    \subsubsection{Evaluation on Single-Traversal Scenarios}
    We conduct quantitative evaluations on the Argoverse 2 and Waymo Open datasets. Table~\ref{tab:single_travel_compare} reports performance on two tasks: novel view synthesis (NVS), which assesses generalization to unseen views, and scene reconstruction, which evaluates fitting quality on training views.
    As shown in Table~\ref{tab:single_travel_compare}, ADM-GS achieves the best quantitative results among the compared methods on both datasets. On Argoverse 2, our method attains 30.14 dB PSNR for NVS, improving over the strong baseline 4DGF~\cite{fischer2024dynamic} (29.48 dB) by 0.66 dB. Compared with other recent methods such as Bilateral-Driving~\cite{wang2025unifying} and OmniRe~\cite{chen2025omnire}, ADM-GS also shows clear gains across all three image-quality metrics. On the Waymo Open dataset, ADM-GS achieves 28.91 dB PSNR for NVS, outperforming 4DGF by 1.30 dB, while also obtaining the best SSIM and LPIPS. These results demonstrate the effectiveness of ADM-GS for driving scene reconstruction and novel view synthesis.

    \subsubsection{Evaluation on Multi-Traversal Scenarios}
    We evaluate novel view synthesis performance on multi-traversal scenarios using three sequences from the Argoverse 2 dataset. As shown in Table~\ref{tab:multi_traversal_3seq}, ADM-GS achieves the best PSNR and SSIM among the compared methods, improving over 4DGF~\cite{fischer2024dynamic} by 0.98 dB in PSNR (27.38 dB vs.\ 26.40 dB) and by 0.008 in SSIM (0.788 vs.\ 0.780). These results suggest that explicit appearance decomposition is beneficial under larger cross-traversal illumination changes.
    We also compare with MTGS~\cite{li2025mtgs}. While MTGS obtains a lower LPIPS score, it yields lower PSNR and SSIM than ADM-GS. This indicates a trade-off between perceptual similarity and structural fidelity. In contrast, ADM-GS achieves the best pixel-level and structural reconstruction accuracy while maintaining competitive perceptual quality, showing stronger cross-traversal consistency under varying illumination conditions.
    
    \begin{table}[t]
    \centering
    \caption{Quantitative comparison on multi-traversal reconstruction on the Argoverse 2 dataset. The models are trained on three traversals with varying illumination conditions. Bold indicates the best performance.}
    \label{tab:multi_traversal_3seq}
    \setlength{\tabcolsep}{12pt} 
    \begin{tabular}{lccc}
    \toprule
    \textbf{Method} & \textbf{PSNR} $\uparrow$ & \textbf{SSIM} $\uparrow$ & \textbf{LPIPS} $\downarrow$ \\
    \midrule
    4DGF~\cite{fischer2024dynamic} & \underline{26.40} & \underline{0.780} & 0.388 \\
    MTGS~\cite{li2025mtgs} & 23.28 & 0.717 & \textbf{0.287} \\
    \textbf{ADM-GS (Ours)} & \textbf{27.38} & \textbf{0.788} & \underline{0.368} \\
    \bottomrule
    \end{tabular}
    \end{table}
    
\subsection{Analysis of Appearance Decomposition}
\label{sec:decomposition_analysis}

\begin{figure*}[t]
    \centering
    \offinterlineskip 
    
    \newcommand{\SideLabelW}{0.025\linewidth}
    \newcommand{\ImgW}{0.193\linewidth}
    \newcommand{\HGap}{\hspace{1pt}}
    
    \newcommand{\CImg}[1]{%
        \raisebox{-0.5\height}{\includegraphics[width=\ImgW]{#1}}%
    }
    \newcommand{\CLabel}[1]{%
        \raisebox{-0.5\height}{%
            \makebox[\SideLabelW][c]{\rotatebox{90}{\footnotesize \textbf{#1}}}%
        }%
    }
    \newcommand{\CHeader}[1]{%
        \makebox[\ImgW][c]{\small \textbf{#1}}%
    }

    \makebox[\SideLabelW]{}%
    \HGap
    \CHeader{Rendered RGB}%
    \HGap
    \CHeader{Material (Invariant)}%
    \HGap
    \CHeader{Illumination (Varying)}%
    \HGap
    \CHeader{Normal}%
    \HGap
    \CHeader{Depth}%
    
    \vspace{3pt} \par 

    \CLabel{Traversal A}%
    \HGap
    \CImg{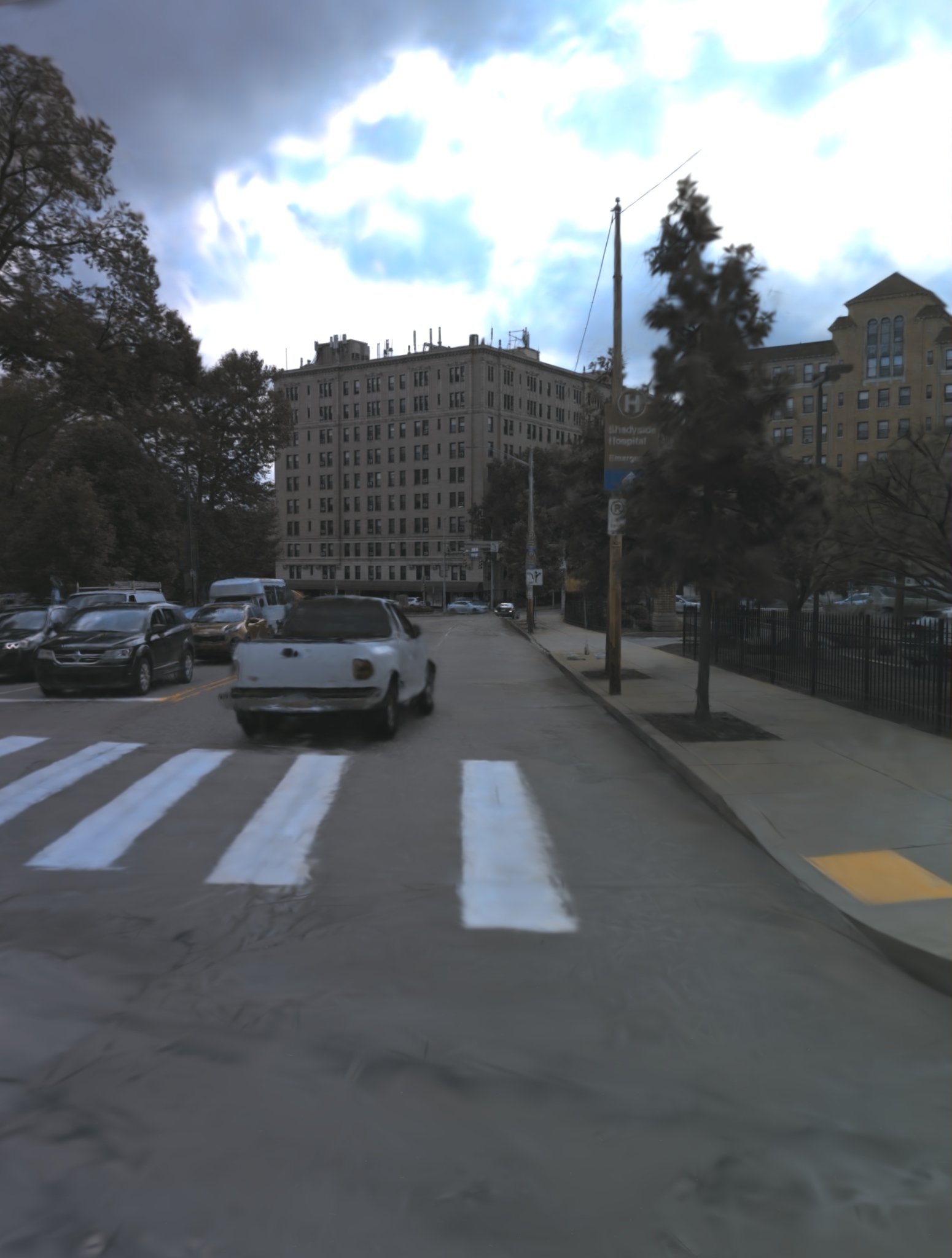}%
    \HGap
    \CImg{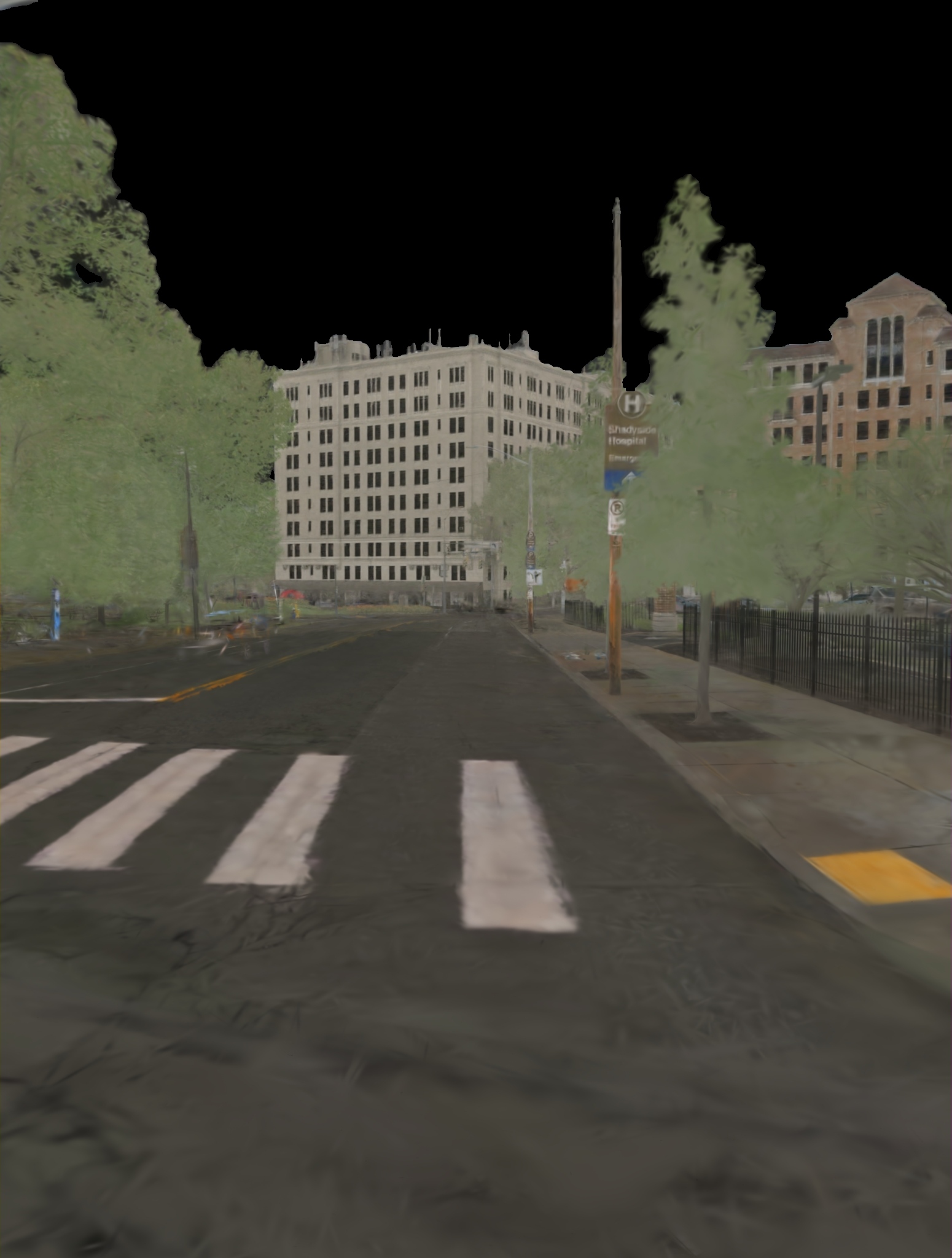}%
    \HGap
    \CImg{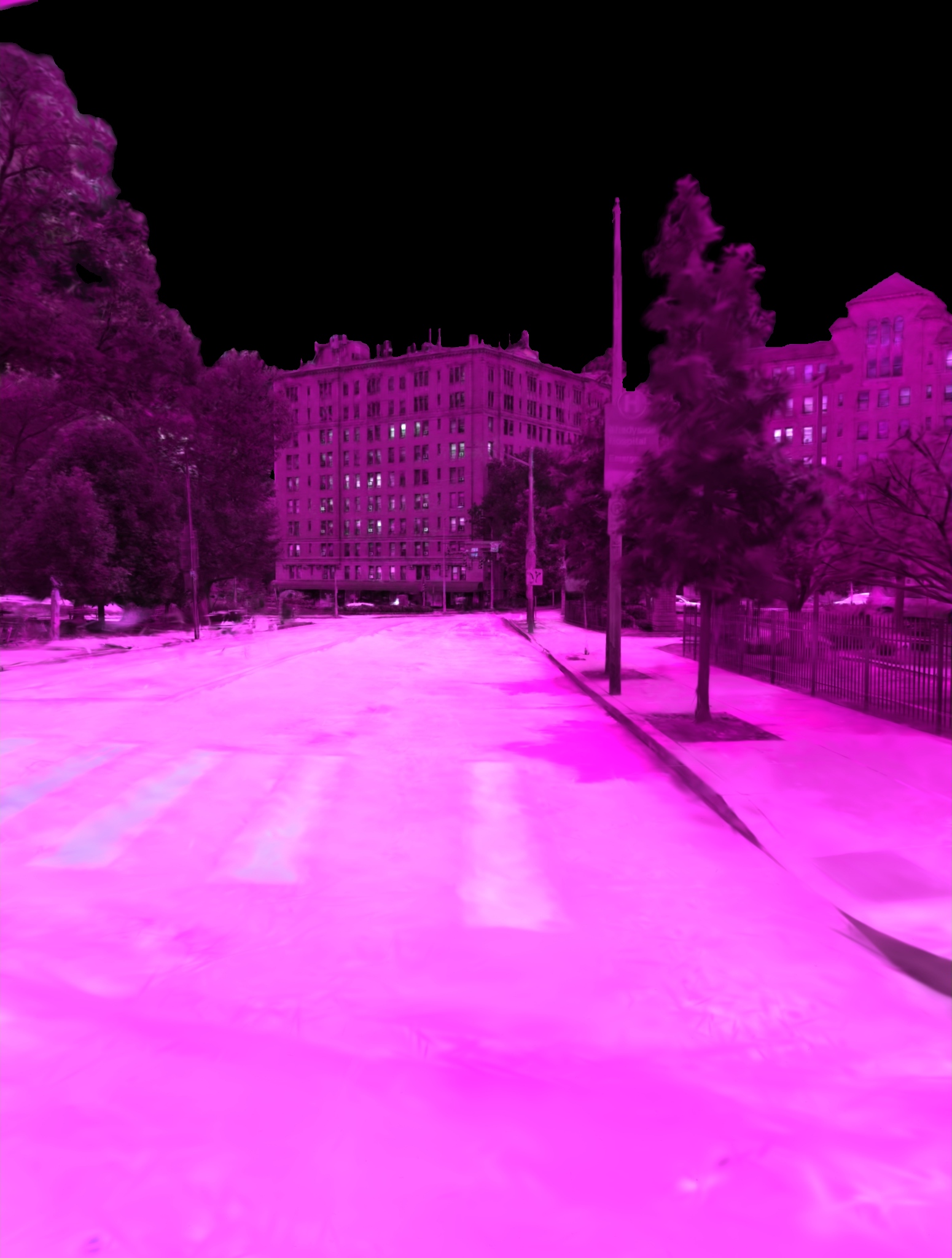}%
    \HGap
    \CImg{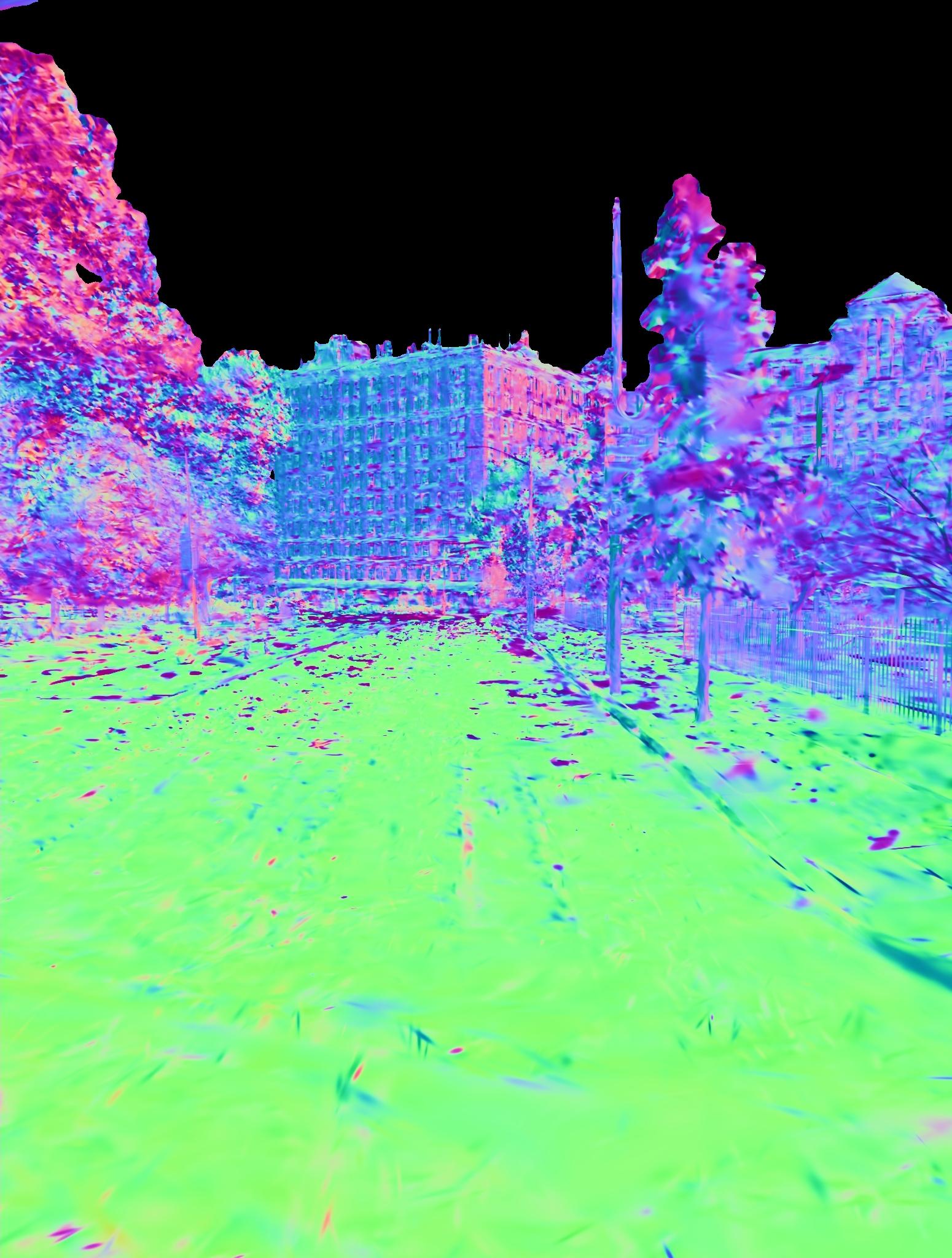}%
    \HGap
    \CImg{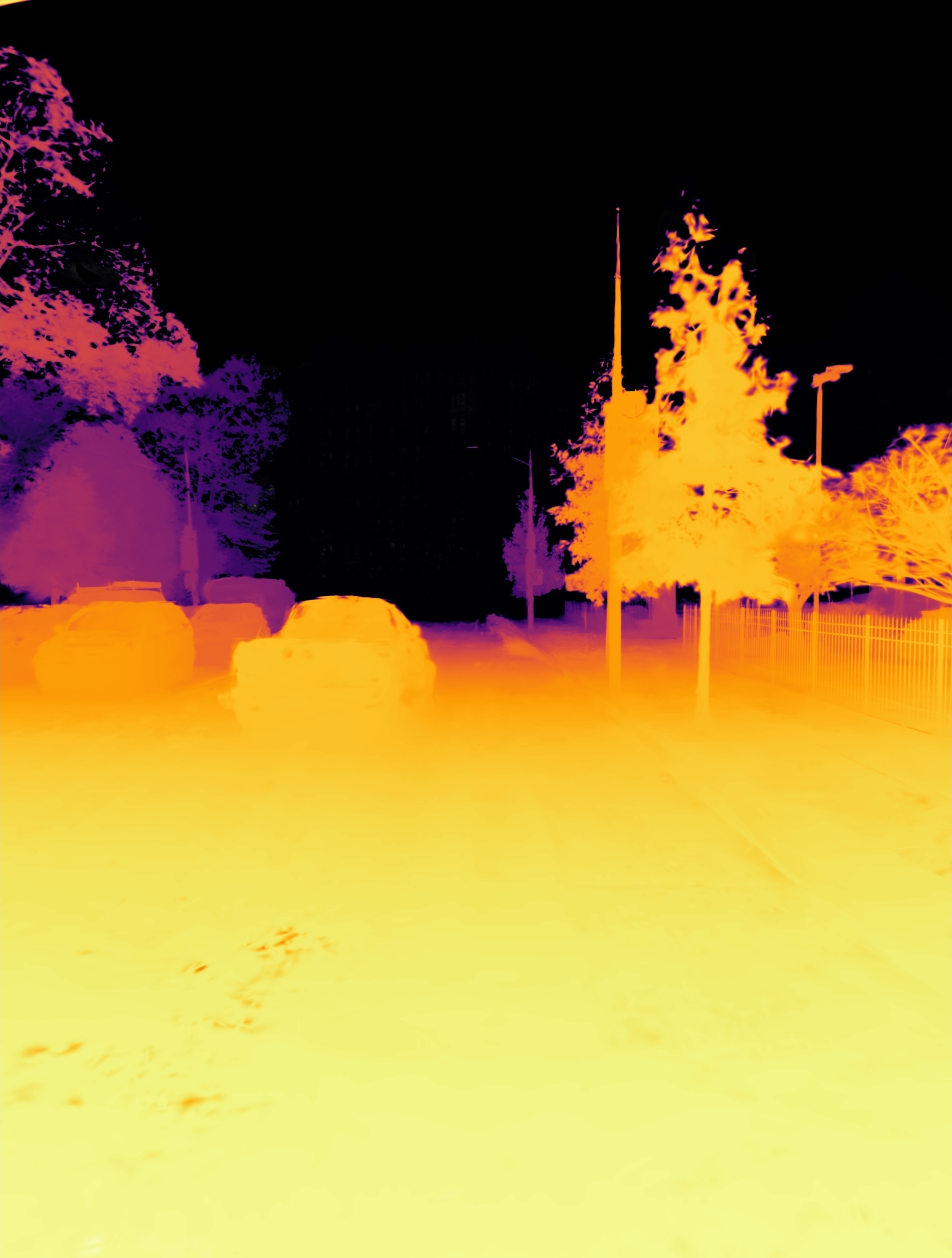}%
    
    \vspace{1pt} \par 

    \CLabel{Traversal B}%
    \HGap
    \CImg{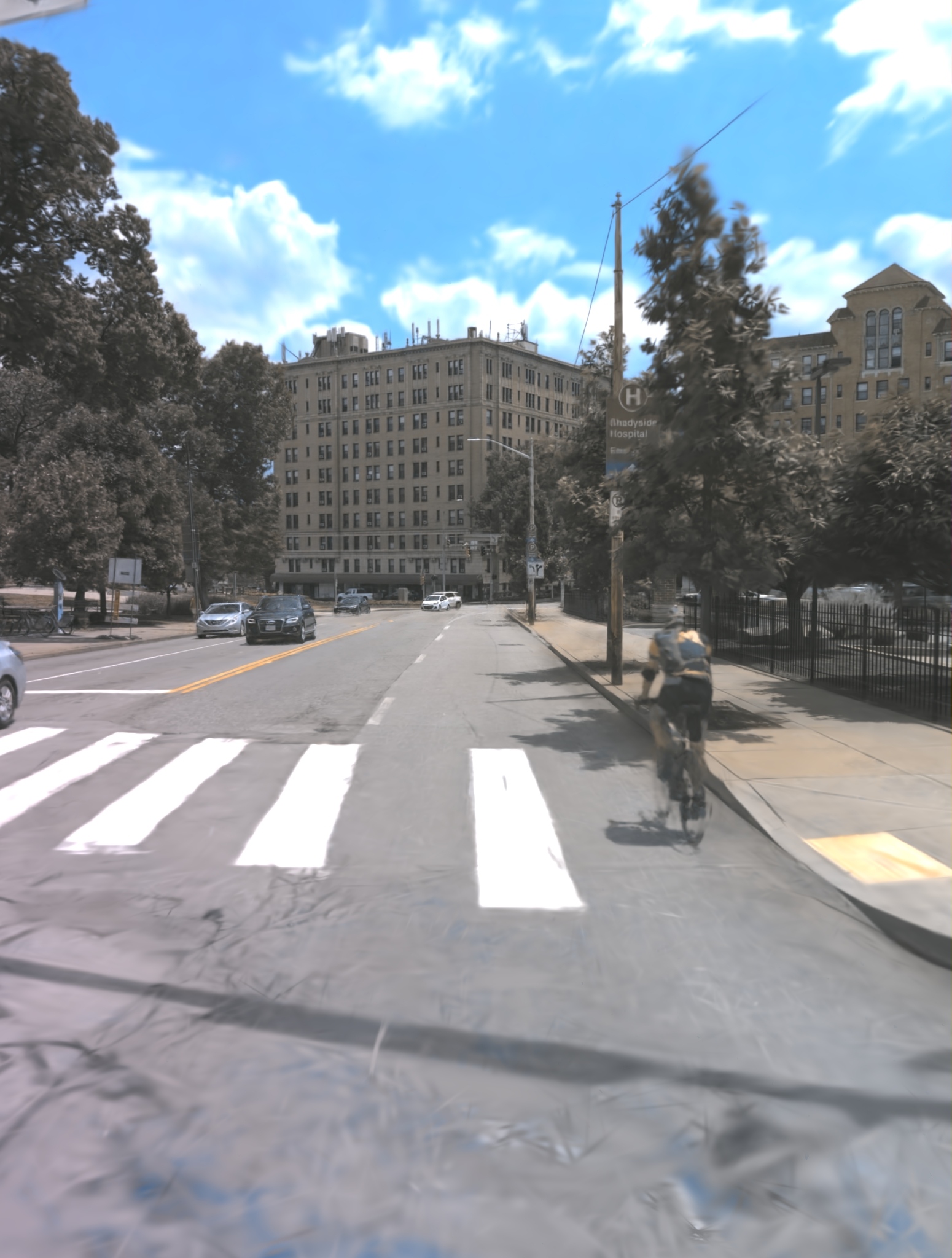}%
    \HGap
    \CImg{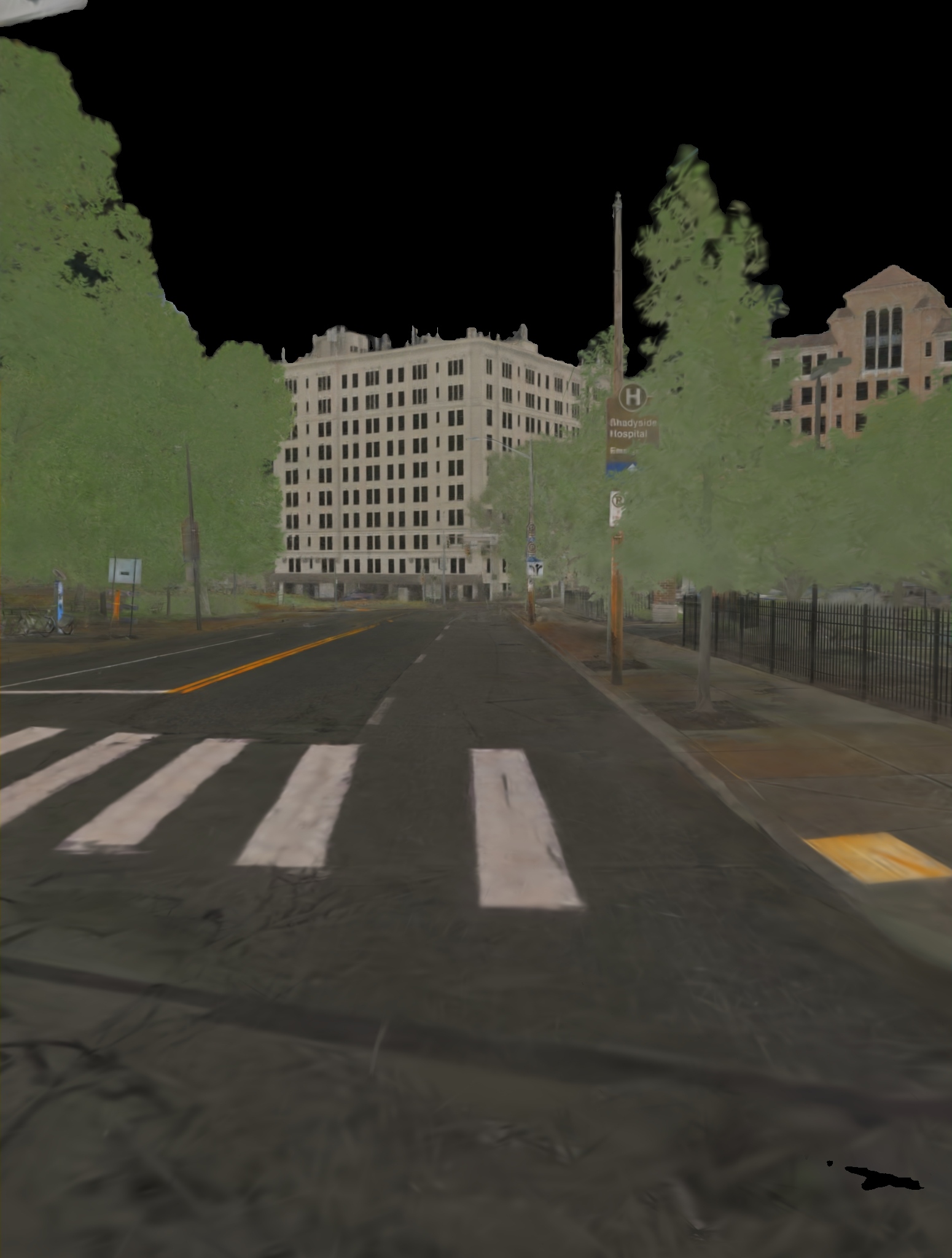}%
    \HGap
    \CImg{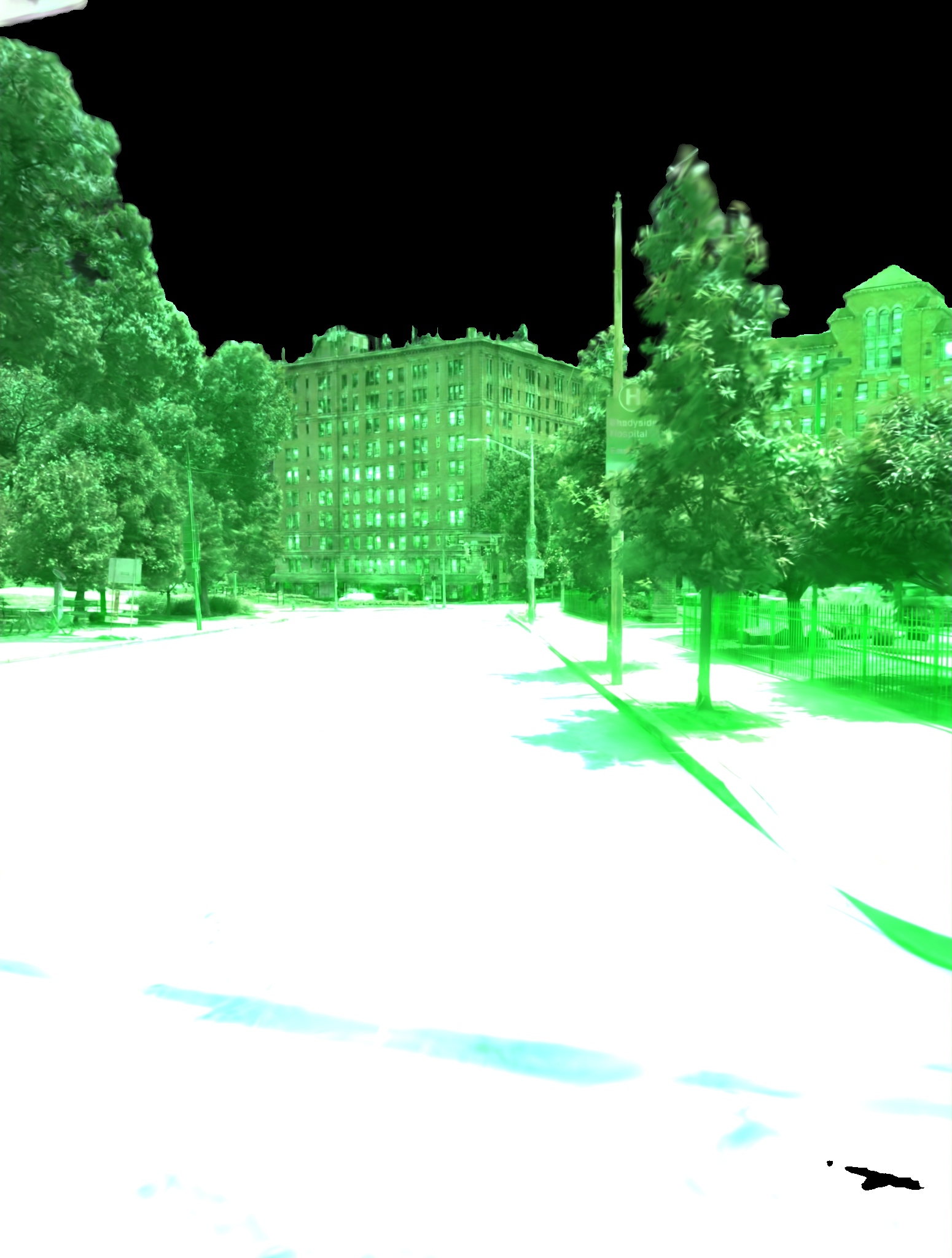}%
    \HGap
    \CImg{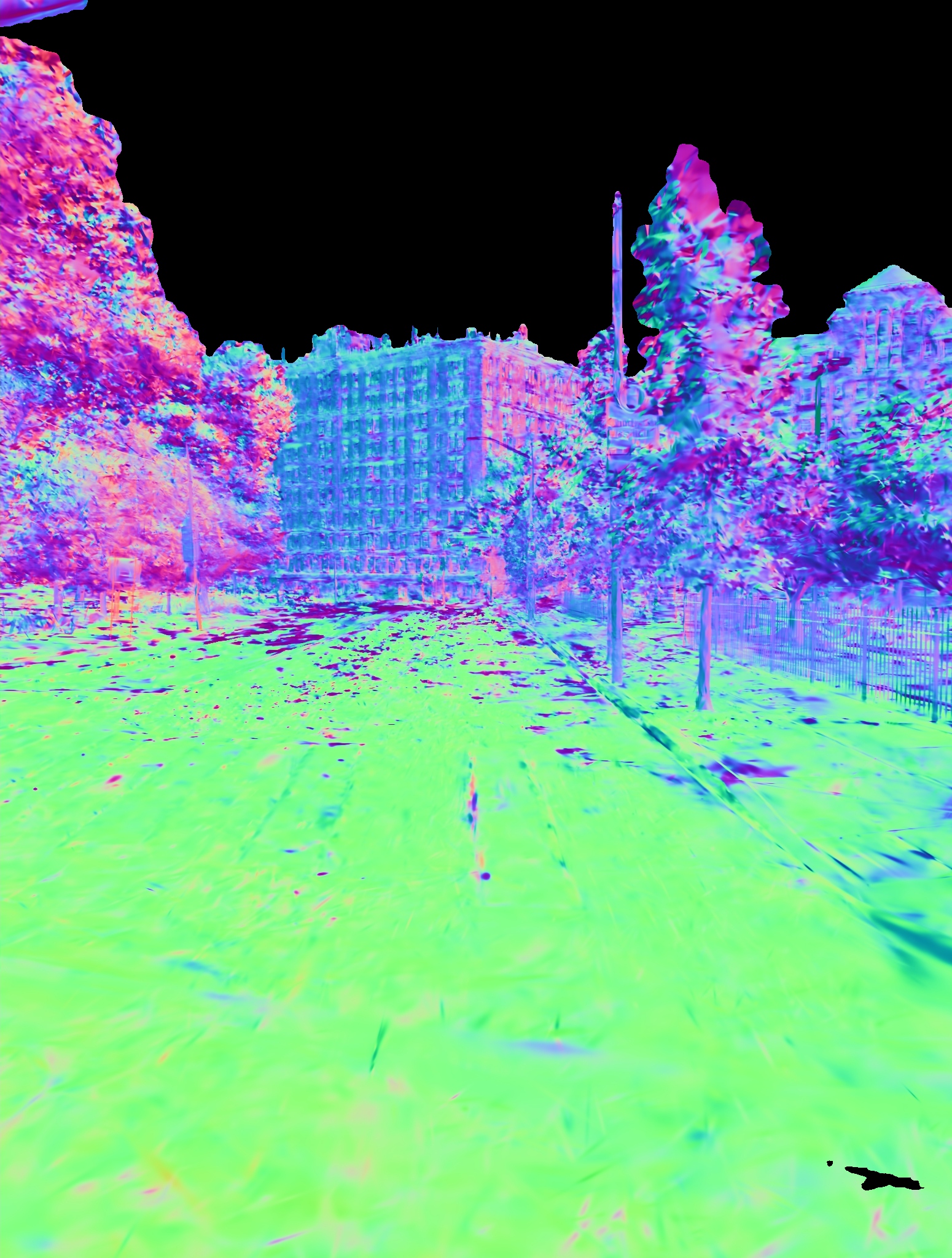}%
    \HGap
    \CImg{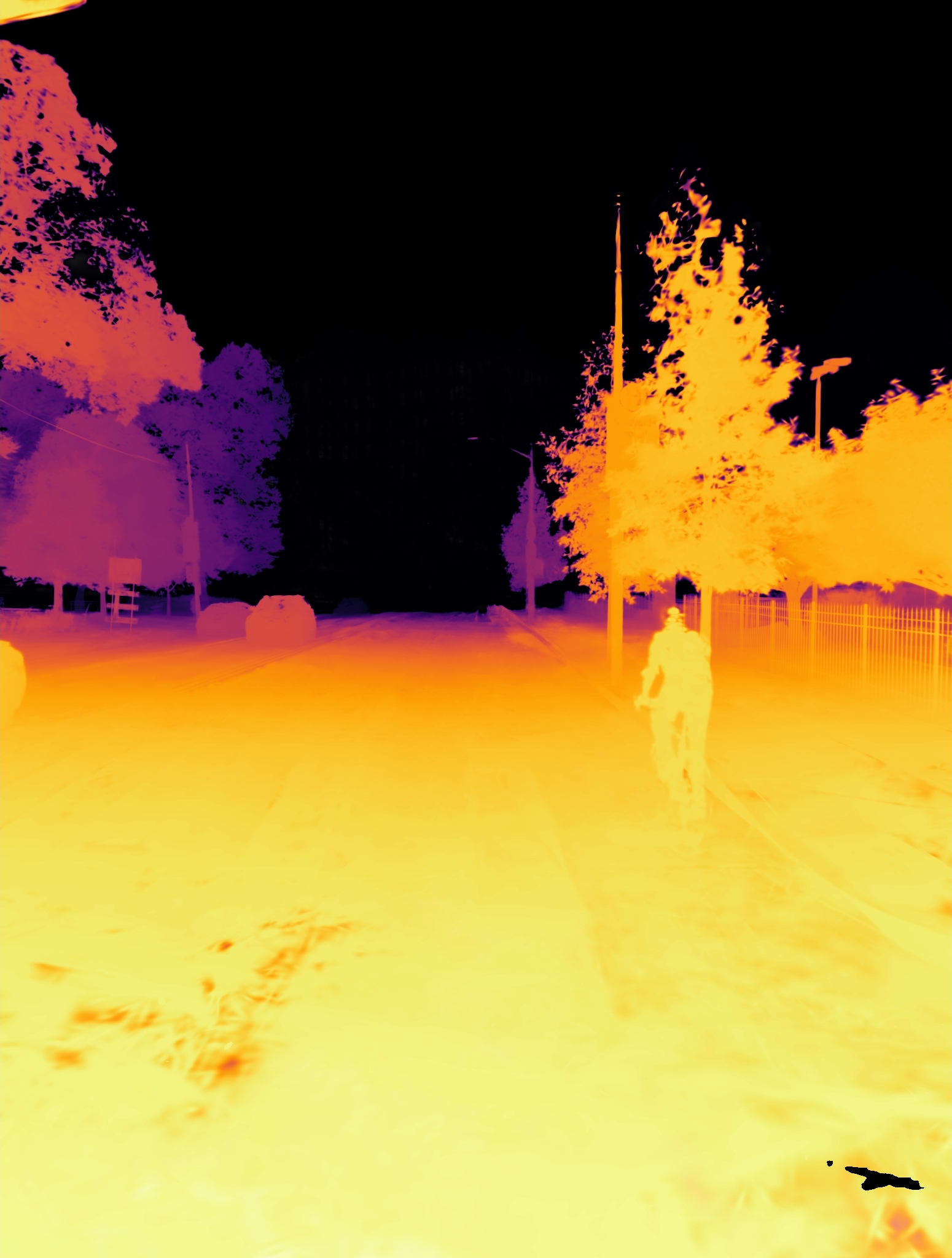}%

    \vspace{5pt}
    \caption{\textbf{Qualitative analysis of appearance decomposition on the Argoverse 2 multi-traversal dataset.} Each row corresponds to a different traversal of the same scene. The estimated material remains largely consistent across traversals, while the illumination maps capture traversal-dependent illumination changes. The normal maps provide geometric cues that support illumination estimation.}
    \label{fig:decomposition_vis}
\end{figure*}

To analyze the decomposition learned by ADM-GS, we visualize the estimated appearance components in Fig.~\ref{fig:decomposition_vis}. We select two traversals of the same scene captured at different times to illustrate the cross-traversal consistency of the decomposition.

\vspace{0.3em}
\textbf{Material consistency and illumination variation:}
As shown in the second and third columns of Fig.~\ref{fig:decomposition_vis}, the decomposition separates the scene appearance into traversal-invariant material and traversal-dependent illumination. On dominant structural surfaces, the estimated material maps remain consistent across traversals, while the illumination maps reflect changes in illumination, cast shadows, and view-dependent highlights.

\vspace{0.3em}
\textbf{Geometric support:}
The fourth column shows the surface normals. Explicit 3D Gaussian primitives may still exhibit local granularity, and some texture details can be reflected in the estimated geometry. Nevertheless, these normals are sufficiently stable to guide the light field, suggesting that the learned geometry provides useful structural cues for appearance decomposition.

\begin{figure*}[t]
    \centering
    \offinterlineskip
    \newcommand{\RelightImgW}{0.24\linewidth}
    \newcommand{\RelightGap}{\hspace{1pt}}
    \newcommand{\RHeader}[1]{\makebox[\RelightImgW][c]{\small \textbf{#1}}}
    \newcommand{\RImg}[1]{\includegraphics[width=\RelightImgW]{#1}}

    \RHeader{Source Scene ($T_1$)}%
    \RelightGap
    \RHeader{Target Illumination ($T_2$)}%
    \RelightGap
    \RHeader{Estimated Material ($\hat{\mathbf{M}}_{T_1}$)}%
    \RelightGap
    \RHeader{Relit Result ($T_1 \leftarrow T_2$)}%
    \vspace{2pt} \par

    \RImg{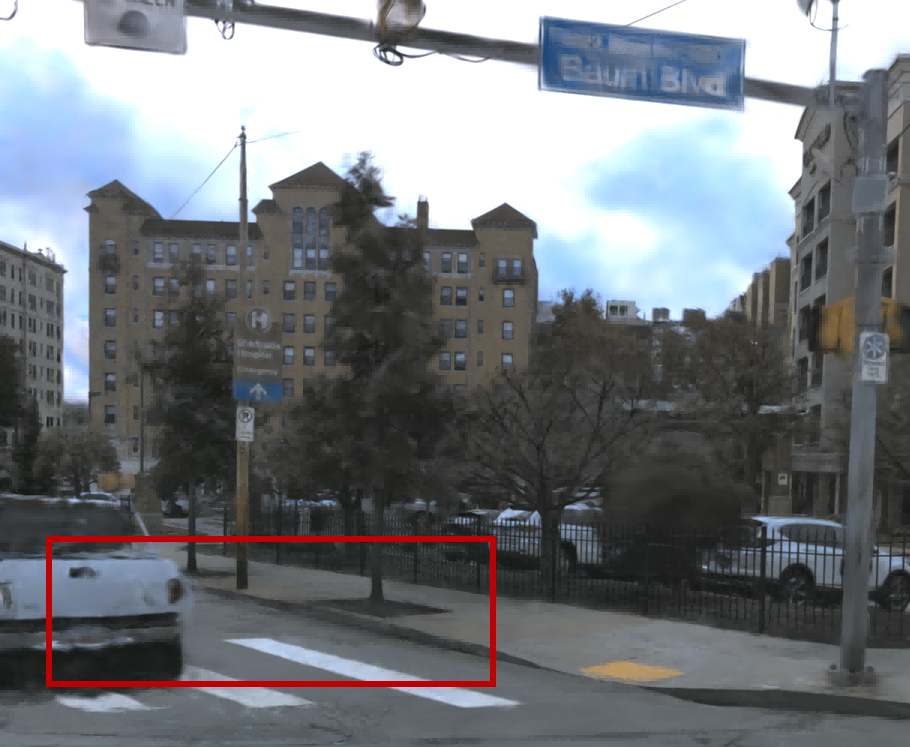}%
    \RelightGap
    \RImg{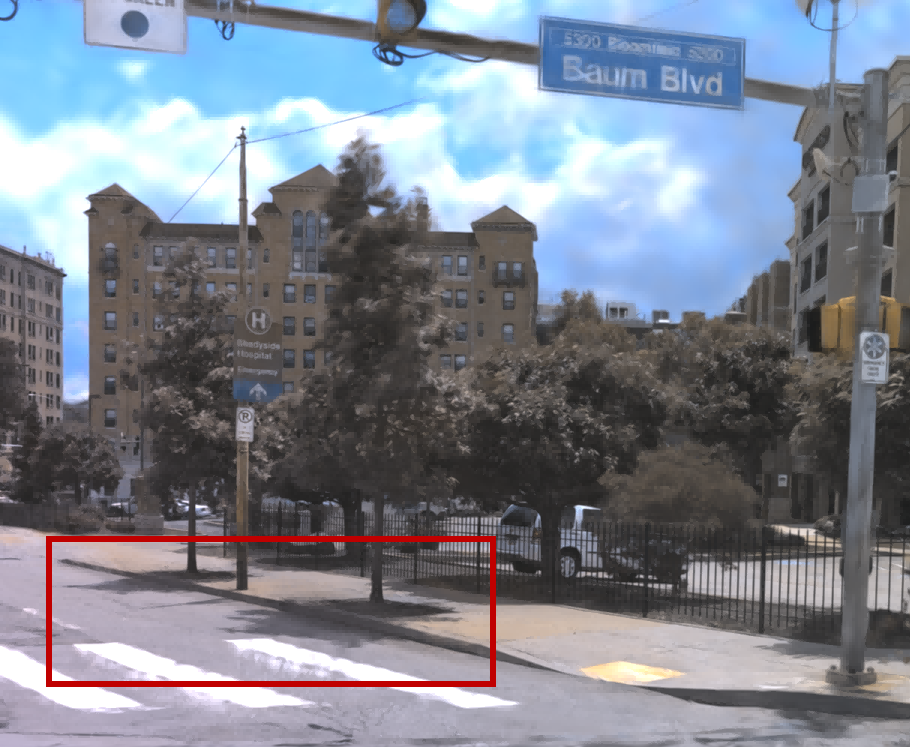}%
    \RelightGap
    \RImg{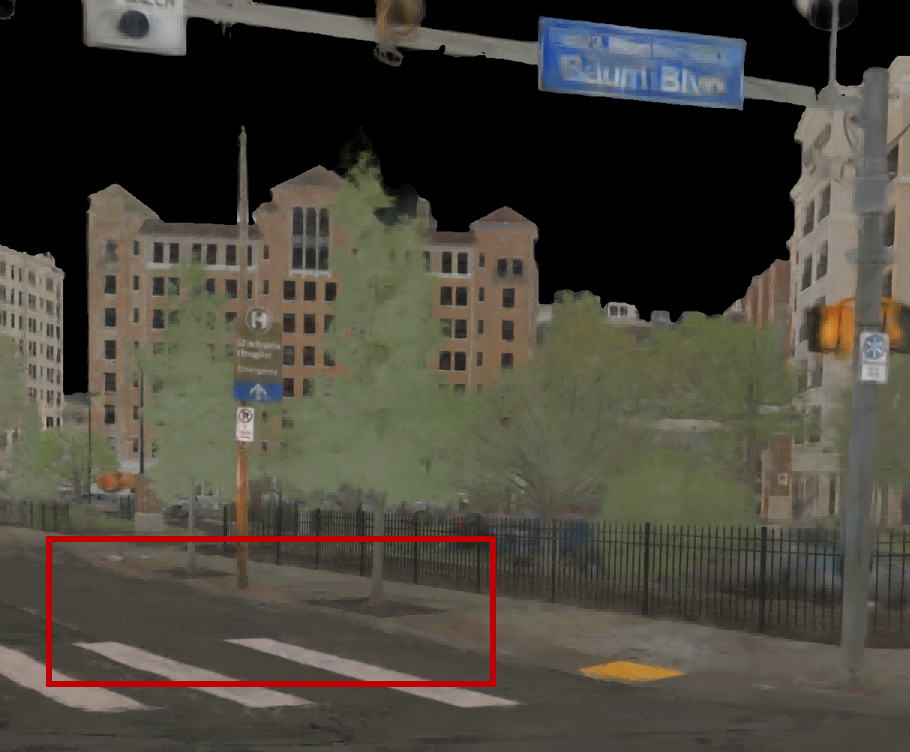}%
    \RelightGap
    \RImg{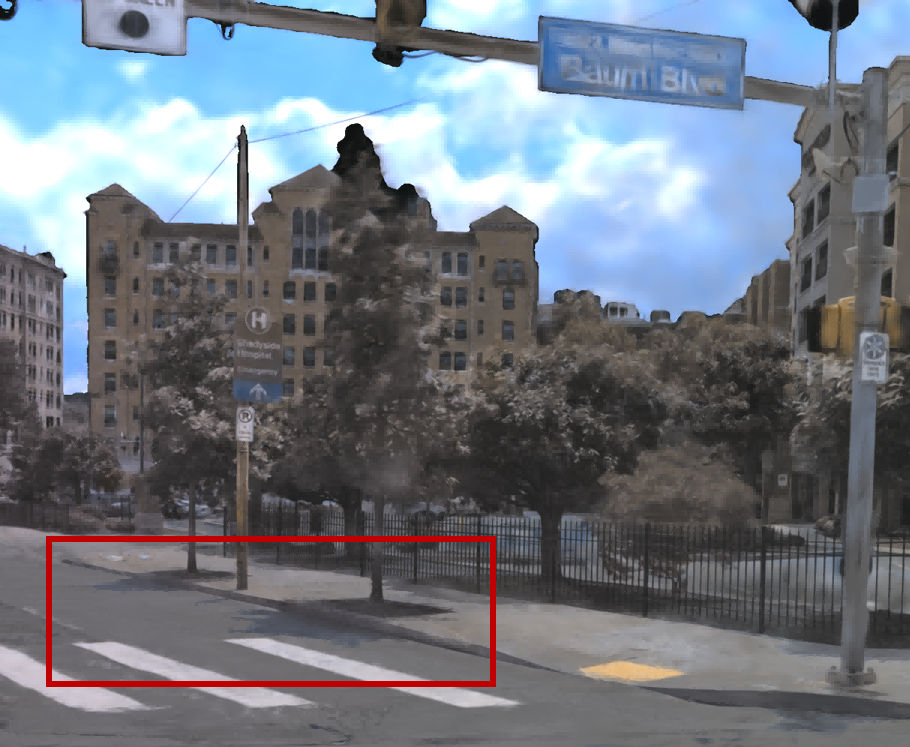}%

    \vspace{5pt}
    \caption{\textbf{Qualitative application: cross-traversal scene relighting.}
    We demonstrate the controllability of the learned decomposition by transferring traversal-dependent illumination from Traversal $T_2$ to a scene observed in Traversal $T_1$.
    Column 1 shows the source image from $T_1$.
    Column 2 shows the target traversal used to provide the illumination condition from $T_2$.
    Column 3 shows the estimated traversal-invariant material map $\hat{\mathbf{M}}_{T_1}$.
    Column 4 shows the relit result obtained by combining the material from $T_1$ with the illumination condition from $T_2$.
    Observe that the shadow cast by the tree (red box) follows the illumination pattern of $T_2$, while the underlying road texture remains consistent with $T_1$.
    The relighting operation is applied only to the static background, while dynamic foreground objects are not edited.}
    \label{fig:relighting}
\end{figure*}

\subsection{Qualitative Application: Scene Relighting}
\label{sec:relighting}

To further illustrate the controllability of the learned decomposition, we conduct a scene relighting experiment. Leveraging the decomposed representation of ADM-GS, we can manipulate scene appearance by combining the estimated material from one traversal with the illumination condition from another traversal. In Fig.~\ref{fig:relighting}, we show a cross-traversal relighting example in which the illumination pattern from a sunny traversal ($T_2$) is transferred to a scene captured under an overcast traversal ($T_1$).

Let $\hat{\mathbf{M}}_{T_1}$ denote the estimated material map from Traversal $T_1$, and let $\mathbf{L}(\mathbf{x},\mathbf{n},\mathbf{v},\mathbf{r},e_{T_2})$ denote the light field conditioned on the traversal embedding of $T_2$. The relit image is synthesized as
\begin{equation}
\hat{\mathbf{I}}^{\mathrm{relit}}_{T_1\leftarrow T_2}
=
\hat{\mathbf{M}}_{T_1}
\odot
\mathbf{L}(\mathbf{x},\mathbf{n},\mathbf{v},\mathbf{r},e_{T_2}).
\end{equation}

As shown in Fig.~\ref{fig:relighting}, the source scene ($T_1$, Col.~1) exhibits diffuse illumination with weak cast shadows, whereas the target traversal ($T_2$, Col.~2) contains stronger directional sunlight and more distinct shadow patterns. By combining the material from $T_1$ with the illumination condition from $T_2$, our method produces a relit result (Col.~4) that follows the target traversal's illumination pattern while preserving the source scene's underlying appearance structure. In particular, the shadow cast by the tree shifts toward the pattern observed in Traversal $T_2$, while the road texture remains consistent with Traversal $T_1$. Note that this relighting operation is applied only to the static background representation and dynamic foreground objects such as vehicles are not relit in this experiment.


\subsection{Ablation Studies }
To validate the design choices of ADM-GS, we conduct ablation studies at two levels: a system-level analysis of the overall appearance decomposition framework and a fine-grained analysis of the illumination MLP design.

    \subsubsection{System-Level Effectiveness} 
    We first evaluate the contribution of the core architectural components on the Argoverse 2 multi-traversal dataset, focusing on the novel view synthesis (NVS) task under varying illumination. Quantitative results are summarized in Table~\ref{tab:system_ablation}.

    \textbf{Implicit vs. Explicit.}
    Starting from a latent appearance baseline, we introduce explicit appearance decomposition together with pseudo material supervision to stabilize the decomposition. This improves PSNR from 27.10 dB to 27.31 dB, suggesting that explicit appearance decomposition is beneficial under appearance variation across traversals.
    
    \textbf{Geometry-aware design.}
    Based on the decomposition model, we further introduce geometry-aware components, including pseudo-normal supervision, normal/reflection cues, and geometric regularization. This further improves reconstruction quality to 27.39 dB PSNR and 0.371 LPIPS, suggesting that geometry-aware constraints improve illumination estimation and appearance decomposition.

    \begin{table}[t]
    \centering
    \caption{\textbf{System-level ablation on Argoverse 2.} We analyze the transition from implicit appearance modeling to explicit appearance decomposition on the multi-traversal NVS task. Explicit Decomposition includes pseudo material supervision, and + Geometry Constraints further adds pseudo normal supervision together with geometric cues.}
    \label{tab:system_ablation}
    \renewcommand{\arraystretch}{1.2}
    \setlength{\tabcolsep}{10pt}
    
    \begin{tabular}{l ccc}
    \toprule
    \textbf{Method Configuration} & \textbf{PSNR} $\uparrow$ & \textbf{SSIM} $\uparrow$ & \textbf{LPIPS} $\downarrow$ \\
    \midrule
    \textbf{Implicit Appearance} & 27.10 & 0.782 & 0.383 \\
    \textbf{Explicit Decomposition} & 27.31 & 0.785 & 0.380 \\
    \textbf{+ Geometry Constraints} & \textbf{27.39} & \textbf{0.789} & \textbf{0.371} \\
    \bottomrule
    \end{tabular}
    \end{table}

    \subsubsection{Fine-Grained illumination MLP Design} 
    To further investigate the role of the illumination design, we conduct a module-level ablation of the illumination MLP on the Waymo Open Dataset. We analyze the impact of geometric supervision ($\mathcal{L}_{normal}$) and different input feature combinations on the unseen test set, as shown in Table~\ref{tab:mlp_ablation}.

    \textbf{Geometric supervision ($\mathcal{L}_{normal}$).}
    By introducing pseudo-normal supervision derived from monocular depth estimation, we obtain more stable surface normals from the Gaussian field. Although these normals are not yet fed into the illumination MLP, this supervision alone improves PSNR from 28.65 to 28.77 dB, suggesting that geometry-aware supervision benefits illumination estimation and rendering quality.
    
    \textbf{Reflection vector ($\mathbf{r}$).}
    Explicitly introducing the reflection vector $\mathbf{r}$ reduces LPIPS from 0.196 to 0.187. Since $\mathbf{r}$ aligns more closely with the dominant direction of the specular lobe, it provides a useful inductive bias for modeling high-frequency view-dependent highlights.
    
    \textbf{Completeness ($\mathbf{r} + \mathbf{n}$).}
    Finally, supplementing the input with surface normals $\mathbf{n}$ yields the best performance (28.91 dB PSNR / 0.186 LPIPS). This suggests that while $\mathbf{r}$ helps capture sharper view-dependent effects, explicitly providing $\mathbf{n}$ offers complementary cues for diffuse illumination.
    
    \begin{table}[t]
    \centering
    \caption{\textbf{Ablation of input features for the illumination MLP on Waymo.} We analyze the impact of geometric supervision ($\mathcal{L}_{normal}$) and different input feature combinations ($r, n$) on the unseen test set.}

    \label{tab:mlp_ablation}
    \renewcommand{\arraystretch}{1.2}
    \setlength{\tabcolsep}{3.5pt} 
    
    \begin{tabular}{l c cc ccc} 
    \toprule
    \multirow{2}{*}{\textbf{Variant}} & \multirow{2}{*}{$\mathcal{L}_{normal}$} & \multicolumn{2}{c}{\textbf{Inputs}} & \multicolumn{3}{c}{\textbf{Metrics}} \\ 
    \cmidrule(lr){3-4} \cmidrule(lr){5-7}
     & & $r$ & $n$ & PSNR $\uparrow$ & SSIM $\uparrow$ & LPIPS $\downarrow$ \\
    \midrule
    Baseline & $\times$ & $\times$ & $\times$ & 28.65 & 0.856 & 0.211 \\
    + Normal Loss & \checkmark & $\times$ & $\times$ & 28.77 & 0.859 & 0.196 \\
    + Reflection & \checkmark & \checkmark & $\times$ & 28.86 & 0.861 & 0.187 \\
    + Refl. \& Normal & \checkmark & \checkmark & \checkmark & \textbf{28.91} & \textbf{0.862} & \textbf{0.186} \\
    \bottomrule
    \end{tabular}
    \end{table}

\section{Conclusion}
This paper presents ADM-GS, a 3D Gaussian Splatting framework for multi-traversal autonomous driving scene reconstruction with explicit appearance decomposition. By introducing traversal-invariant material and traversal-dependent illumination, ADM-GS alleviates appearance inconsistencies across traversals captured under varying illumination conditions. Our illumination design further improves the modeling of both diffuse and view-dependent effects. Quantitative and qualitative evaluations on the Argoverse 2 and Waymo Open datasets demonstrate the effectiveness of ADM-GS, including a +0.98 dB PSNR improvement over a strong baseline in multi-traversal reconstruction.
The decomposition-based representation presented in this work provides a useful direction for consistent urban scene modeling and may benefit future research on outdoor scene editing and relightable digital twin generation. At the same time, our current training still relies on pseudo supervision from monocular material estimation, and reducing this dependency is an important direction for future work. In addition, extending ADM-GS from the current qualitative relighting demonstration to more comprehensive and controllable outdoor relighting remains an open challenge. Future work will explore more explicit illumination modeling, including sun-sky representations and solar irradiance estimation, to improve controllability in complex outdoor environments.


\bibliographystyle{IEEEtran}
\bibliography{IEEEabrv,ref}

\vfill

\end{document}